\documentclass[pdflatex,sn-mathphys-num,twoside,iicol]{sn-jnl}% Math and Physical Sciences Numbered Reference Style

\usepackage{geometry}
\usepackage{graphicx}%
\usepackage{multirow}%
\usepackage{amsmath,amssymb,amsfonts}%
\usepackage{amsthm}%
\usepackage{mathrsfs}%
\usepackage[title]{appendix}%
\usepackage{xcolor}%
\usepackage{textcomp}%
\usepackage{manyfoot}%
\usepackage{booktabs}%
\usepackage{algorithm}%
\usepackage{algorithmicx}%
\usepackage{algpseudocode}%
\usepackage{listings}%
\usepackage{enumitem}

\theoremstyle{thmstyleone}%
\theoremstyle{thmstyletwo}%

\theoremstyle{thmstylethree}%

\begin{document}

\title[Article Title]{UCLIR: Unsupervised Continual Learning using Independent Representations}

%{An Autonomous Training-Aware Agent for Iterative Model Improvement}

%%=============================================================%%
%% GivenName	-> \fnm{Joergen W.}
%% Particle	-> \spfx{van der} -> surname prefix
%% FamilyName	-> \sur{Ploeg}
%% Suffix	-> \sfx{IV}
%% \author*[1,2]{\fnm{Joergen W.} \spfx{van der} \sur{Ploeg} 
%%  \sfx{IV}}\email{iauthor@gmail.com}
%%=============================================================%%

\author[1,2]{\fnm{Kamel} \sur{Aouaidjia}}\email{kamel@henu.edu.cn}

\author[1,2]{\fnm{Yanchao} \sur{Duan}}\email{yanchao.duan@foxmail.com}
\author[1,2]{\fnm{Zihui} \sur{Zhen}}\email{690367242@qq.com}
\author[1,2]{\fnm{Xin} \sur{Li}}\email{782045777@qq.com}
\author[1,2]{\fnm{Yongfei} \sur{Lv}}\email{henulvyongfei@163.com}
%\equalcont{These authors contributed equally to this work.}

%\author[1]{\fnm{Third} \sur{Author}}\email{iiiauthor@gmail.com}
%\equalcont{These authors contributed equally to this work.}

\author[3]{\fnm{Chongsheng} \sur{Zhang}}\email{cszhang@henu.edu.cn}
%\equalcont{These authors contributed equally to this work.}

%School of Information Engineering, Henan Kaifeng College of Science Technology and Communication, 475004 Kaifeng, China.
\affil[1]{\orgdiv{School of Information Engineering}, \orgname{Henan Kaifeng College of Science Technology and Communication}, \orgaddress{ \city{Kaifeng}, \postcode{475000}, \state{Henan}, \country{China}}}

\affil[2]{\orgname{Henan Engineering Research Center for AI Technology of Converged Media}, \orgaddress{ \city{Kaifeng}, \postcode{475000}, \state{Henan}, \country{China}}}

\affil[3]{\orgdiv{School of Computer and
Information Engineering}, \orgname{Henan University}, \city{Kaifeng}, \postcode{475001}, \state{Henan}, \country{China}}

%\affil[3]{\orgdiv{Department}, \orgname{Organization}, \orgaddress{\street{Street}, \city{City}, \postcode{610101}, \state{State}, \country{Country}}}

%%==================================%%
%% Sample for unstructured abstract %%
%%==================================%%

\abstract{
In continual learning, regularization-based methods aim to mimic  neuroplasticity by updating network weights to learn new tasks while constraining changes to parameters that are important for previously learned tasks through penalty terms. However, as the number of tasks increases, preserving knowledge from earlier tasks becomes less effective. This is mainly because updates for new tasks are distributed across the entire network, where parameters are highly interdependent due to the dense connectivity between neurons. Inspired by the Exponential Distance Rule (EDR), which states that the number of neural connections decreases with increasing physical distance, learning is naturally locally correlated while inter-area updates remain less task-dependent. Motivated by this principle, in this paper we demonstrate that applying weight regularization to multiple small and independently learned complementary representations is more effective than regularizing a single large shared representation. By decomposing the input into independently learned components, learning new tasks induces only localized parameter updates within specific representation, which limit interference and improve knowledge retention across tasks. We employ independent autoencoders that learn different parts of the input independently through unsupervised regularization.  At inference time,  a linear classifier is used for evaluation. The experimental results on a sequence of tasks using 5-Splits MNIST, sequence of 5-datasets, 5-Splits CIFAR-10, 10-Splits Tiny ImageNet shows competitive results with supervised and replay-based methods. Code is available$^{1}$.}


%%\pacs[JEL Classification]{D8, H51}

%%\pacs[MSC Classification]{35A01, 65L10, 65L12, 65L20, 65L70}

\maketitle

\section{Introduction}\label{sec1}

Recent advances in artificial intelligence have been largely driven by the availability of massive training datasets. However, regardless of data scale, modern deep models exhibit limited ability to generalize to unseen or unfamiliar scenarios due to the lack of dynamic learning mechanisms that enable continuous adaptation to new data. Since these models are typically trained in an offline manner using static datasets, their reasoning capability remains constrained by the diversity of the original training data. These limitations stem from architectural designs optimized for single-task learning with large datasets processed all at once, lacking the ability to continually learn new representations from a sequence of tasks without eliminating the previous learned representations. The inability of neural networks to preserve previously acquired knowledge while learning new tasks is known as catastrophic forgetting. This motivated the development of continual learning frameworks, which are  generally categorized into replay-based \cite{shin2017continual}, architectural expansion \cite{rusu2016progressive}, regularization-based approaches \cite{kirkpatrick2017overcoming}, or hybrid methods \cite{yoon2018lifelong}. 

The human biological learning system operates by adapting synaptic strength between neurons or forming new connections when acquiring new knowledge \cite{hebb1949organization,bi_synaptic_1998-1}. This  principle lines up with regularization approaches that propose techniques to preserve previous knowledge by carefully updating the weights of the new task without significantly changing the weights of the neurons that contributed to the previous tasks \cite{ahn2019uncertainty,chen2021overcoming}. Although such techniques have shown improvement in continue learning, they have the following limitations: (1) Their performance starts to decline as the number of tasks increases (2) Connections between neurons  are strongly interdependence; therefore, during regularization, changing an unimportant weight in the current task can influence an important weight. (3) When all tasks are encoded within a single latent space, task-specific features become highly correlated. This correlation causes interference and reduces the model’s ability to preserve diverse task knowledge over long task sequences.

\footnotetext[1]{\href{https://github.com/adjkamel/UCLIR}{\url{https://github.com/adjkamel/UCLIR}}}

The main reasons for the previous limitations is that current methods use a single shared representation to learn new tasks,  lacking a dynamic learning mechanism that learns independent, complementary representations which can be regularized independently, reducing inter-dependencies between neurons. This assumption is inspired by  the Exponential Distance Rule (EDR) \cite{ercsey2013predictive} observed in the brain, which suggests that neural connections weaken as physical distance increases, leading to the hypothesis that learning primarily occurs through local synaptic updates, while interactions between distant regions are sparse and task-dependent. For example, in the visual cortex, different regions are specialized for processing features at different levels of abstraction, with areas responding to low-level features such as edges, and other areas encoding higher-level features such as shapes \cite{kobatake1994neuronal}. However, in current continual learning models, for a given task such as image classification with CNNs, low-level features across different classes are strongly dependent on shared high-level representations. As a result, updating low-level layers often requires modifying later layers as well, which reduces the flexibility to learn new task-specific low-level representations without influencing existing high-level representations.

Based on the previous limitations and analysis, we propose a relay-free unsupervised representation continual learning framework that uses multiple independent complementary representations to learn different parts of the same input, where the regularization is performed on each representation independently using an autoencoder that reconstructs the input image. We adopt two methods of independent representations: Independent hierarchical representations and independent parallel representations. In the former, three independent autoencoders (LevelNets) process the same input and each extracts a different level of features. In the latter, five similar independent autoencoders (SideNets), each process a cropped parts of the input and reconstructs it independently. Within each task and each representation, weight regularization is applied to preserve the important weights of the previous tasks while learning a new task. At inference time, predictions are obtained using a linear classifier trained on top of the frozen representations.

The contributions of this work can be summarized in the following points:

\begin{figure*}
\centering
\includegraphics[width=1\linewidth]{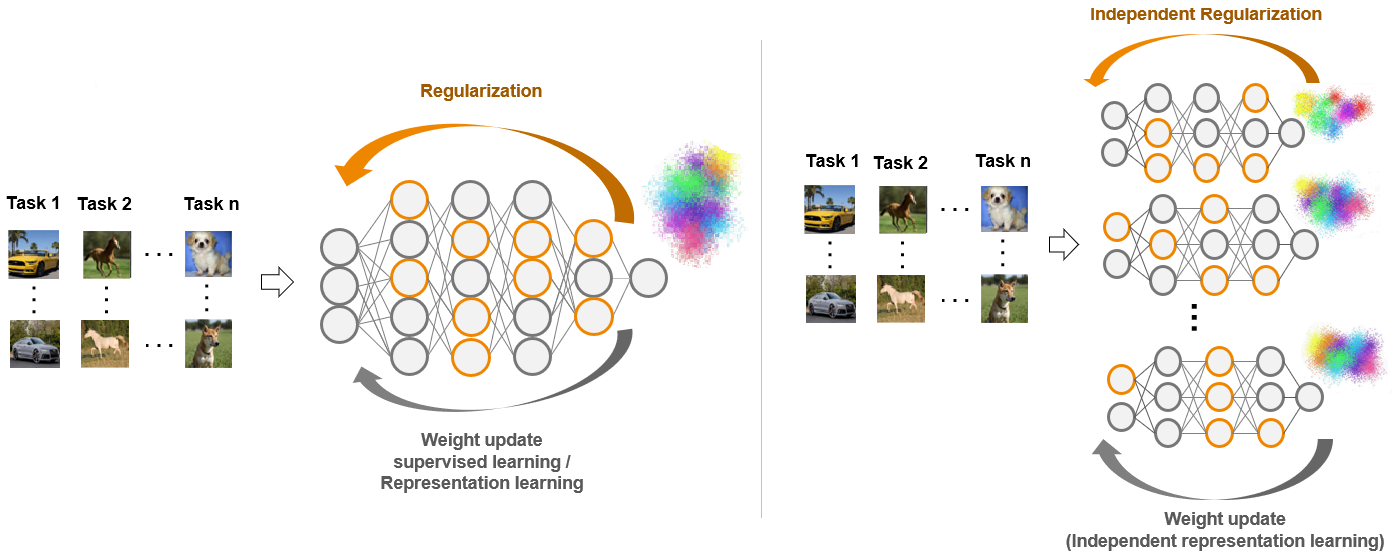}
\caption{Overall illustration of the proposed continual learning framework using independent regularized representations (right), compared to existing methods with a single representation (left). Colored nodes represent neurons that are important for previous tasks.}
\label{fig:main} 
\end{figure*}

\begin{figure*}[t]
\centering
\includegraphics[width=0.8\linewidth]{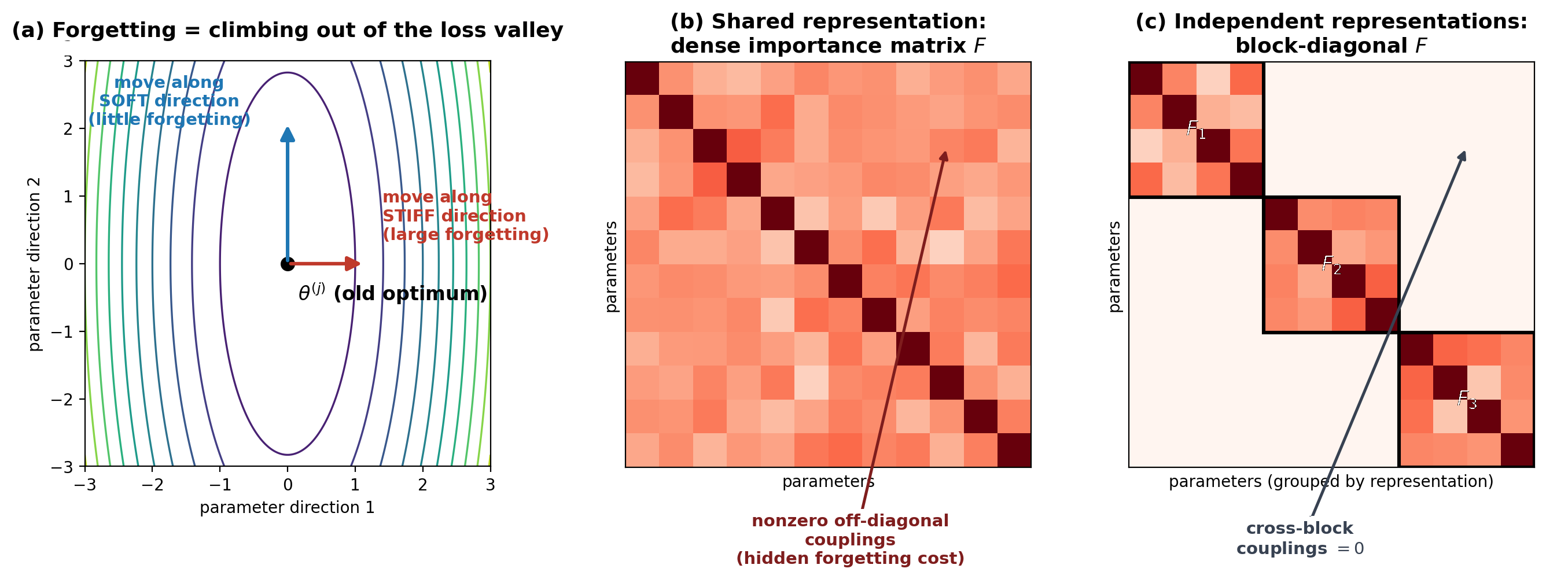}
\caption{Geometric intuition for the theoretical analysis. \textbf{(a)} After learning task $j$, the weights  $\theta^{(j)}$ sit at a minimum of the loss. Learning a later task moves the weights away from it and increases the loss, which is forgetting. A small move along a stiff direction (high curvature) causes a large loss increase, while even a large move along a soft direction (low curvature) causes little. \textbf{(b)} For a single shared representation, the importance matrix $F$ is dense: the off-diagonal entries are the hidden couplings that diagonal EWC ignores. \textbf{(c)} For independent representations, $F$ is block-diagonal ($F_1,\dots,F_R$). The cross-block couplings are exactly zero, so forgetting separates into one term per representation and each block can be protected on its own.}
\label{fig:explain}
\end{figure*}

\begin{itemize}

\item We propose a replay-free unsupervised continual learning framework that replaces regularization over a single shared representation with regularization across multiple small, independent complementary representations, each specializing in a distinct part of the input. By localizing task-specific updates within these independent representations, our approach effectively mitigates harmful interference as the number of tasks grows.

\item We instantiate the independent representation framework as two CNN
architectures: depth-wise independent representations, which extract
complementary features across multiple representational levels via varying
receptive fields, and space-wise independent representations, which
extract complementary features from distinct spatial crops using identically
structured branches.

\item Evaluation of the proposed method on a sequence of continual learning tasks using 5-Splits MNIST, CIFAR-10, 10-Splits Tiny ImageNet, and 5-datasets demonstrate its effectiveness. Despite being unsupervised and replay-free, our method achieves competitive performance compared with supervised and replay-based methods.

\end{itemize}

\section{Related Work}

Regularization-based methods constrain parameter updates to preserve important weights. Elastic Weight Consolidation (EWC) estimates parameter importance using Fisher information and penalizes changes to critical parameters \cite{kirkpatrick2017overcoming}. Synaptic Intelligence accumulates path-based importance measures to regulate online plasticity \cite{zenke2017continual}. Uncertainty-aware regularization reduces the number of parameters required for implementing per-weight regularization using node-wise uncertainty \cite{ahn2019uncertainty}, while Bayesian generative regularization introduces probabilistic priors to stabilize continual updates \cite{chen2021overcoming}. Replay-based methods reduce forgetting by using stored or generated samples from previous tasks. Strong experience replay distills the knowledge of past experiences stored in the memory buffer to preserve backward consistency, and further enforces forward consistency by mimicking the old model’s responses on current training data \cite{shin2017continual}, while generative replay methods retain knowledge by training a generative model to generate samples of the past data instead of using raw data \cite{lopez2017gradient}. Adversarial continual learning incorporates adversarial objectives to preserve data distributions across tasks \cite{ebrahimi2020adversarial}. Architectural approaches dynamically expand or reorganize networks to allocate capacity for new tasks. Progressive neural architecture grows the model structure as new tasks arrive \cite{rusu2016progressive}, while increasing-depth strategies add new layers over time to isolate task-specific knowledge \cite{kozal2023increasing}. 

Self-supervised contrastive learning methods  \cite{zbontar2021barlow}, \cite{chen2020simple} are used to mitigate forgetting through continual representation learning.  \citet{cha2021co2l} proposed to learn representations using contrastive learning, and preserve learned representations using self-supervised distillation. \citet{fini2022self} show how self-supervised loss functions can  be converted into a distillation mechanism that is used to map the current state of the representation to the past state. \citet{wen2024provable} also adopt the idea of using knowledge distillation with contrastive loss by providing a theoretical explanation of how the pretraining can benefit continual learning. Moreover, they proposed an adaptive distillation that assigns coefficients for different tasks. \citet{cheng2023cucl} introduced a Codebook for feature representation to increase the discriminative ability in the continual learning process, then a contrastive loss is used on both the original representations and the Codebook features. Recent works have also explored learning without reliance on stored past information.  \citet{cheng2024must} proposed an information-free method that is based on a self-supervised pretext task called Alternate Task Discrimination to improve anti-forgetting capability via encouraging the model to discriminate which data stream current sample is from using contrastive learning on augmented data.

In contrast to the previous methods, our approach is replay-free and does not rely on contrastive learning, pretrained models or distillation. Instead, it only uses independently trained, regularized representations for image reconstruction.

\begin{figure*}
\centering
\includegraphics[width=1\linewidth]{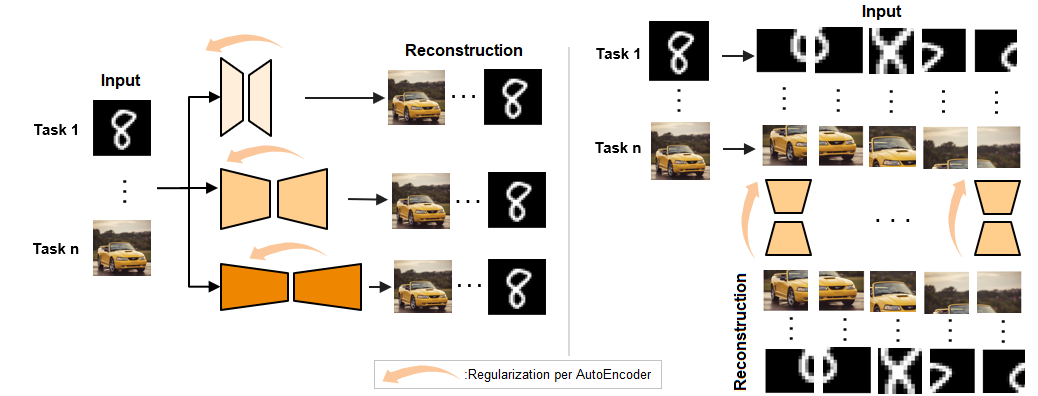}
\caption{Left: Independent hierarchical representations (LevelNets). Right: Independent parallel representations (SideNets).}
\label{fig:splits} 
\end{figure*}

\section{Method}

\subsection{Theoretical Analysis}
\label{sec:theory}

Fig.~\ref{fig:main} illustrates the proposed idea compared to previous single-model regularization. In this section we explain why regularizing several small, independently learned representations is preferable to regularizing one shared representation of the same total size. Fig.~\ref{fig:explain} summarizes this argument geometrically. We first describe what forgetting is and why it can be written as a simple weighted cost. We then show that making the representations independent removes the harmful cross-terms responsible for interference, and we relate this to the two proposed architectures, the depth-wise (LevelNets) and space-wise (SideNets) partitions, in the next section.

\subsubsection{Forgetting as a weighted cost of parameter shift}
\label{sec:explain}

Consider a model with parameters $\theta \in \mathbb{R}^{d}$, with $d$ the total number of parameters in the model, trained one task at a time on $T_1,\dots,T_K$. Let $\theta^{(j)}$ be the weights right after task $j$ has been learned well, and let them shift to $\theta^{(k)}$ after a later task $k$ is learned. The shift in the weights caused by this later learning is:

\begin{equation}
\Delta = \theta^{(k)} - \theta^{(j)}
\label{eq:th-delta}
\end{equation}

Learning task $k$ shift the weights, and moving weights that mattered for task $j$ makes task $j$ worse. This loss of past performance is forgetting. To quantify it, we need a notion of how important each weight is for task $j$, i.e.\ how strongly task $j$'s loss reacts when that weight is nudged. This is captured by the Fisher information matrix $F^{(j)}$:

\begin{equation}
F^{(j)} =
\mathbb{E}_{x \sim D_j}
\!\left[
\nabla_\theta \log p(x\mid\theta)\,
\nabla_\theta \log p(x\mid\theta)^{\top}
\right]_{\theta=\theta^{(j)}}
\label{eq:th-fisher}
\end{equation}

\noindent Whose diagonal entry $F^{(j)}_{ii}$ measures the importance of weight $i$ on its own, and whose off-diagonal entry $F^{(j)}_{im}$ measures how much weights $i$ and $m$ matter \emph{together}. Formally, $F^{(j)}$ is the curvature of the loss around the optimum $\theta^{(j)}$. Since $\theta^{(j)}$ is a minimum, the loss increases in every direction as the parameters shift away from it, at a rate set by the local curvature (Fig.~\ref{fig:explain}(a)). High curvature (large importance) causes a large loss increase for a given parameter change, whereas low curvature causes only a small loss. In Fig.~\ref{fig:explain}(a) this is the difference between the stiff direction, where a small shift already raises the loss sharply and causes large forgetting, and the soft direction, where even a large shift raises it only slightly and costs little. This increase in loss is the forgetting of task $j$, and it is given by the weighted cost:

\begin{equation}
\mathcal{F}_{j\to k}
\;\triangleq\;
\mathcal{L}^{(j)}_{\mathrm{rec}}(\theta^{(k)})
-\mathcal{L}^{(j)}_{\mathrm{rec}}(\theta^{(j)})
\;\approx\;
\tfrac{1}{2}\,\Delta^{\top} F^{(j)} \Delta 
\label{eq:th-forget}
\end{equation}

\noindent The total forgetting is how far each weight shifted, weighted by how important it was, summed over all weights and all pairs of weights. Writing this sum out separates two contributions:

\begin{equation}
\mathcal{F}_{j\to k}
\approx
\underbrace{\tfrac{1}{2}\sum_{i} F^{(j)}_{ii}\,\Delta_i^{2}}_{\text{each weight alone}}
\;+\;
\underbrace{\tfrac{1}{2}\sum_{i\neq m} F^{(j)}_{im}\,\Delta_i \Delta_m}_{\text{pairs of weights together}} 
\label{eq:th-forget-entry}
\end{equation}

\noindent The first sum is exactly the term that per-parameter EWC penalizes, through $\tfrac{\lambda}{2}\sum_i F^{(k-1)}_{i}(\theta^{(k)}_i-\theta^{(k-1)}_i)^2$ in Eq.~\eqref{eg:1}. The second sum is a hidden cost, where even a tiny shift $\Delta_i$ in a seemingly unimportant weight can still cause forgetting if that weight is coupled ($F^{(j)}_{im}$ large) to an important weight $m$. These couplings are the off-diagonal entries of the dense importance matrix in Fig.~\ref{fig:explain}(b). Diagonal EWC ignores this second sum entirely, so in a densely connected network it misses a real source of interference. This is the geometric statement of limitation~(2) in the introduction.

\subsubsection{Eliminate the hidden cost through independent representations}
\label{sec:theory-independence}

Now split the weights into $R$ separate groups (blocks), one per representation:
\begin{equation}
\theta = [\theta_1^{\top},\theta_2^{\top},\dots,\theta_R^{\top}]^{\top},
\qquad \theta_r \in \mathbb{R}^{d_r},\quad \sum_r d_r = d
\label{eq:th-partition}
\end{equation}
Where each block $\theta_r$ is a separate autoencoder $Enc_r$ that sees its own view $v_r(x)$ of the input and is trained by its own reconstruction loss $\mathcal{L}^{(k)}_{r}$. Because block $r$ is trained only on its own view, its output depends only on its own weights $\theta_r$ and not on any weight in another block. The joint importance of two weights in different blocks is therefore zero:

\begin{equation}
\begin{aligned}
F^{(j)}_{im} &= \mathbb{E}\!\left[
\frac{\partial \log p_r}{\partial \theta_i}
\frac{\partial \log p_{r'}}{\partial \theta_m}
\right] = 0,\\
&\quad i\in \text{block } r,\; m\in \text{block } r',\; r\neq r'
\end{aligned}
\label{eq:th-blockdiag}
\end{equation}

The importance matrix is nonzero only inside each block and zero between blocks, as shown
in Fig.~\ref{fig:explain}(c) and in Eq.~\eqref{eq:th-forget}. It splits into a clean sum, one term per representation, with no cross-representation
interaction left:

\begin{equation}
\mathcal{F}_{j\to k}
\approx
\sum_{r=1}^{R}
\tfrac{1}{2}\,\Delta_r^{\top} F^{(j)}_{r}\,\Delta_r ,
\qquad \Delta_r = \theta_r^{(k)}-\theta_r^{(j)} 
\label{eq:th-forget-sep}
\end{equation}

The interfering pairs that connected different representations in Eq.~\eqref{eq:th-forget-entry}, i.e.\ the off-diagonal mass of Fig.~\ref{fig:explain}(b), are now exactly zero, leaving only the within-block entries of Fig.~\ref{fig:explain}(c). Because each block is small (its size $d_r \ll d$) these
remaining pairs are limited too. Applying EWC to each autoencoder separately therefore controls almost all of the surviving damage, and the error from ignoring the few within-block pairs shrinks as the blocks get smaller. This is the precise sense in which protecting several small independent representations is more
efficient than diagonally protecting one large shared one.

Finally, we bound the worst case. For each block $r$ the most it can forget depends on two quantities. The first is its highest curvature $\rho_r = \lambda_{\max}(F^{(j)}_r)$, the largest eigenvalue of the block's importance matrix $F^{(j)}_r$, which gives the direction in which a parameter
change raises the loss the most. The second is the size of its parameter displacement $\|\Delta_r\|$, the amount block $r$ had to shift to learn task $k$.
From Eq.~\eqref{eq:th-forget-sep} the total forgetting is bounded by:

\begin{equation}
\mathcal{F}_{j\to k}
\;\le\;
\tfrac{1}{2}\sum_{r=1}^{R} \rho_r \,\|\Delta_r\|^{2}
\label{eq:th-bound}
\end{equation}

For a single shared representation of the same total size, the same bound uses one global highest curvature $\rho_{\mathrm{shared}}= \lambda_{\max}(F^{(j)})$ and the
full displacement $\|\Delta\|$, giving $\mathcal{F}_{j\to k}\le \tfrac{1}{2}\rho_{\mathrm{shared}}\|\Delta\|^2$. Two facts make the split version smaller. First, the shared network's highest curvature is at least as large as that of any single block, so $\rho_{\mathrm{shared}}\ge \max_r \rho_r$, where $\max_r \rho_r$ denotes the largest curvature among all blocks. This holds because a larger set of parameters has a larger worst case curvature. Second, each block shifts only a little because its view is a simpler reconstruction target, so $\|\Delta_r\|$ is small. In a shared network every new task keeps updating the same high-curvature direction, so forgetting accumulates as tasks arrive. In the split network each task updates only the block that matches its view and leaves the rest unchanged. This formalizes limitation~(1), that the shared model degrades faster as the number of tasks grows.

\subsection{Independent Representation Decompositions}
We investigate two model-structure approaches for learning independent, complementary representations from the same input. In the first approach, we employ independent hierarchical representations that learn features at three different levels of abstraction using three CNN autoencoders of varying depths, which we refer to as LevelNets. In the second approach, we design independent parallel representations using five CNN autoencoders to extract features from different spatial regions of the input image, which we refer to as SideNets.

Let $T_1,T_2,\dots,T_k$ be a sequence of tasks, where the data of task $k$ is $D_k = \{(x_i^k, y_i^k)\}_{i=1}^{n_k}$. Each task is encoded by $r$ autoencoders that reconstruct the input, producing the $r$ representations $Z_{1k}, Z_{2k}, \dots, Z_{rk}$.

\subsubsection{LevelNets}
In CNNs, early layers typically extract low-level features such as edges and corners, middle layers capture mid-level structures such as shapes, and deeper layers learn high-level, object-level representations. Instead of learning these multi-level features within a single shared hierarchy, we propose to learn them independently by designing three encoders of different depths.  The three autoencoders share the same number of layers; however, they differ in their convolutional filter sizes. In the first autoencoder, small $3\times3$ filters are used in all layers to focus on low-level features. The second autoencoder employs larger $5\times5$ filters to capture higher-level, more abstract features. In the third autoencoder, $7\times7$ filters are used to extract even higher-level semantic information. In all three cases, the encoders take the same input image, and each corresponding decoder reconstructs the image based on the features learned at its respective level throughout the sequence of tasks. This multilevel representation can be extended to include additional encoders with different filter sizes and network depths. In general, the hierarchical feature extraction process can be formulated as: 
\begin{equation}
Z_{lk}=Enc_{lk}(D_k)
\end{equation}

where $l={1,2,\dots, L}$ indexes the encoder, each corresponding to a specific level of feature abstraction. Fig. \ref{fig:splits} (Left) shows an overall illustration of LevelNets approach.

\begin{figure}
\centering
\includegraphics[width=1\linewidth]{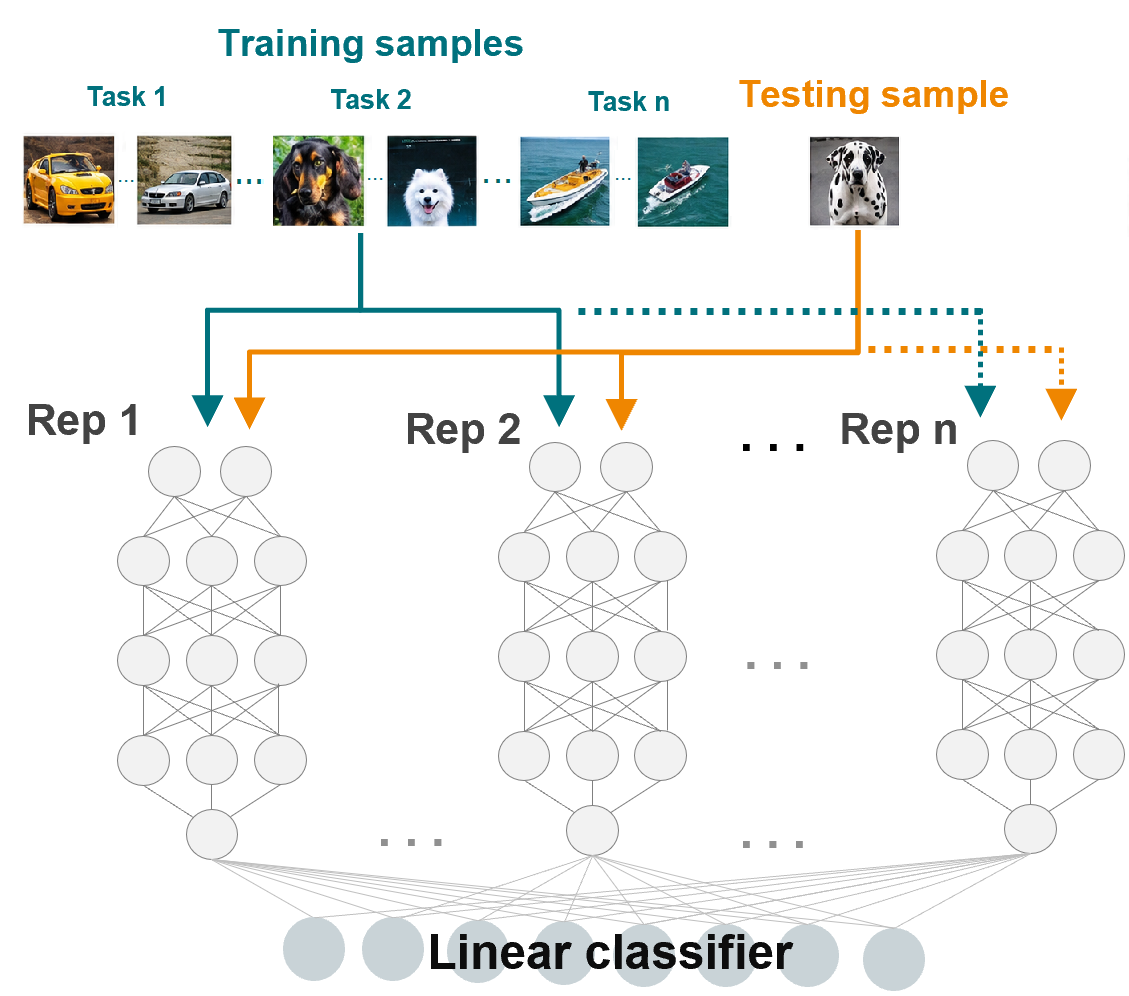}
\caption{Multi-representation inference using a linear classifier trained on the concatenated frozen representations.}
\label{fig:lin} 
\end{figure}

\subsubsection{SideNets}
In the second approach, complementary features are learned across different spatial regions of the input. Specifically, we employ multiple independent autoencoders of equal depth to learn representations from distinct regions of the image. We divide the image into five regions: top-left, top-right, bottom-left, bottom-right, and center. Each autoencoder independently learns features sequentially from one region through reconstruction. All autoencoders share the same architecture, with identical number of layers and identical $3\times3$ convolutional filters. As with the multilevel representations, the number of parallel representations is not restricted to these five regions; fewer or more spatial partitions can be used.  Fig. \ref{fig:splits} (Right) shows an illustration of SideNets approach. The parallel feature extraction process can be formulated as:
\begin{equation}
Z_{sk}=Enc_{sk}(D_k)
\end{equation}
where $s={1, 2,\dots, S}$ indexes the spatial region of each encoder.

\subsubsection{Continual representation learning}
Elastic Weight Consolidation (EWC) \cite{kirkpatrick2017overcoming} introduces constraints during training to prevent significant changes to parameters that are important for previously learned tasks, while learning a new task. In our framework, each representation is learned by a separate autoencoder and updated independently across tasks. We apply EWC to each autoencoder individually, enabling local weight regularization within each representation, without coupling it to the parameters of other representations. This independent regularization strategy reduces inter-representation interference and enhances continual learning performance by distributing task-specific updates across complementary and independent parameter subsets. The reconstruction loss including regularization is formulated as follows:

\begin{equation}
\mathcal{L}_{l}^{(k)} =
\mathcal{L}_{\mathrm{rec}}^{(k)}(Z_{l,k})
+ \frac{\lambda}{2}
\sum_{i}
F_{l,i}^{(k-1)}
\left(
\theta_{l,i}^{(k)} - \theta_{l,i}^{(k-1)}
\right)^2
\label{eg:1}
\end{equation}

where $\mathcal{L}_{\mathrm{rec}}^{(k)}$ is the reconstruction MSE loss for task $k$, $\theta_{l,i}^{(k)}$ denotes the $i$-th parameter of encoder $l$ while learning task $k$, $\theta_{l,i}^{(k-1)}$ are the parameters learned from previous tasks, $F_{l,i}^{(k-1)}$ is the Fisher information estimating the importance of parameter $i$ for encoder $l$, and $\lambda$ controls the strength of the regularization.

\begin{equation}
F_{l,i}^{(k-1)} =
\mathbb{E}_{x \sim D_{k-1}}
\left[
\left(
\frac{\partial \log p(x \mid \theta_l)}{\partial \theta_{l,i}}
\right)^2
\right]
\end{equation}

After training, the learned representations are used for classification without further updating. A linear classifier is trained on top of the frozen representations to assess their discriminative capability (Figure \ref{fig:lin}). In case of LevelNets, features of the three encoders are concatenated to train a linear classifier for inference. Similarly, in the case of SideNets, the classifier is trained on top of the concatenated features of the five encoders.

\section{Experiments}

\subsection{Settings}

We evaluate our method on four continual learning datasets. 5-Splits MNIST partitions the MNIST handwritten digit dataset into five sequential tasks, each containing two classes. 5-Datasets is a sequence of five datasets:  MNIST, notMNIST, SVHN, Fashion-MNIST, and CIFAR-10, where each task includes all ten classes of one dataset. 5-Splits CIFAR-10 divides CIFAR-10 into five sequential tasks with two classes per task. 10-Splits Tiny ImageNet splits the Tiny ImageNet dataset (200 classes) into ten tasks, each containing 20 classes.

Both LevelNets and SideNets are evaluated using a linear classifier trained on their learned representations. For comparison, we report classification accuracy under  class-incremental Learning (CIL) settings over 5 runs of random seeds. For other evaluation experiments, we report task-incremental Learning (TIL) accuracy alongside CIL.

For both LevelNets and SideNets, each independent representation is trained for 50 epochs with the Adam optimizer at a learning rate of 0.001. The EWC regularization strength is fixed to $\lambda = 2$ (Eq. \ref{eg:1}) across all experiments. Linear classifiers are trained for 150 epochs using Adam with a learning rate of 0.0001.

%%%%%%%%%%%%%%%%%%%%%%%%%%%%%%%%%%%  results reported based on adding +- to the main accuracy for fair comparison

\begin{table*}
\caption{CIL accuracy comparison on 5-Splits MNIST.} 
\centering
\small
\setlength{\tabcolsep}{3pt} 
\begin{tabular}{l  l l  c }
\hline
Method & Supervision & Type &  ACC.(\%)\\ 
 \hline
EWC\cite{kirkpatrick2017overcoming}  & Supervised & Regularization & 20.01 $\pm$ 0.06 \\
SI\cite{zenke2017continual}         & Supervised & Regularization & 19.99 $\pm$ 0.06 \\
MAS\cite{aljundi2018memory}         & Supervised & Regularization & 19.52 $\pm$ 0.29 \\
LwF\cite{li2017learning}            & Supervised & Distillation & 24.17 $\pm$ 0.33 \\
GEM\cite{lopez2017gradient}         & Supervised & Replay & 92.20 $\pm$ 0.12 \\
DGR\cite{shin2017continual}         & Supervised & Generative replay & 91.24 $\pm$ 0.33 \\
iCARL\cite{rebuffi2017icarl}        & Supervised & Replay + distillation & 94.57 $\pm$ 0.11 \\
CURL\cite{rao2019continual}         & Unsupervised & Generative replay & 92.59 $\pm$ 0.66 \\
UCLIR (LevelNets)      & Unsupervised & Replay-free  & 97.74 $\pm$ 0.92  \\
UCLIR (SideNets)	  & Unsupervised & Replay-free & \textbf{97.76 $\pm$ 0.57}  \\  %%% 96.02 (not avg)
\hline
\end{tabular}
\label{table:mnist}
\end{table*}

\begin{table*}
\caption{CIL accuracy comparison on 5-Splits CIFAR-10. The compared methods use $200$-example buffer.} 
\centering
\small
\setlength{\tabcolsep}{3pt} 
\begin{tabular}{l  l l  c }
\hline
Method & Supervision & Type & ACC.(\%) \\
\hline
ER \cite{riemerlearning}            & Supervised & Replay & 44.79 $\pm$ 1.86 \\
GEM\cite{lopez2017gradient}             & Supervised & Replay & 25.54 $\pm$ 0.76 \\
A-GEM\cite{chaudhry2019efficient}       & Supervised & Replay & 20.04 $\pm$ 0.34 \\
iCARL \cite{rebuffi2017icarl}            & Supervised & Replay + distillation & 49.02 $\pm$ 3.20 \\
FDR \cite{benjaminmeasuring}         & Supervised & Replay & 30.91 $\pm$ 2.74 \\
GSS \cite{aljundi2019gradient}           & Supervised & Replay & 39.07 $\pm$ 5.59 \\
HAL \cite{chaudhry2021using}             & Supervised & Replay & 32.36 $\pm$ 2.70 \\
UCLIR (LevelNets) 	  & Unsupervised & Replay-free &   \textbf{61.23 $\pm$ 9.56} \\
UCLIR (SideNets)	  & Unsupervised & Replay-free &  54.93 $\pm$ 1.86 \\  %41.09 (not avg)
%%%%%%%%%%% CIL average: 43.86
\hline
\end{tabular}
\label{table:cifar10}
\end{table*}

\subsection{Comparison}

Table~\ref{table:mnist} groups the baselines by supervision and mechanism, spanning supervised regularization (EWC, SI, MAS), distillation (LwF), replay and generative replay (GEM, DGR, iCaRL), and the unsupervised generative-replay method CURL. In the task-incremental setting every method performs well, and UCLIR (SideNets) is competitive at $99.29\%$, essentially tied with the strongest supervised baselines (LwF at $99.60\%$, MAS at $99.22\%$). In
class-incremental learning,  UCLIR (SideNets) reaches the highest class-incremental accuracy ($97.76\%$), exceeding even the supervised replay
methods despite learning its representations without labels and without storing exemplars. The comparison against CURL is the most direct, as it is the
only unsupervised baseline. UCLIR improves on it by over $5\%$ in the class-incremental setting ($97.76\%$ vs.\ $92.59\%$) which indicates that the independent representations capture more class-discriminative structure than CURL's generative approach.

Table~\ref{table:cifar10} compares UCLIR against supervised replay-based baselines on CIFAR-10, all of which rely on a stored $200$-example buffer. In
the class-incremental setting UCLIR (LevelNets) attains $61.23\%$ and UCLIR (SideNets) $54.93\%$, both surpassing the best supervised replay method, iCaRL ($49.02\%$). This holds despite UCLIR using neither labels nor stored exemplars. For task-incremental learning,  UCLIR achieved $80.85\%$ for LevelNets and  $78.60\%$ for SideNets compared to the replay baselines (ER at $91.19\%$, FDR at $91.01\%$, GEM at $90.44\%$). LevelNets achieves the higher mean in both settings but with large variance ($\pm 9.56$ CIL, $\pm 6.38$ TIL), whereas SideNets is far more stable in accuracy over different runs ($\pm 1.86$ and $\pm 0.39$).

Table~\ref{table:tiny} presents the comparison on the longer, 10-Splits Tiny ImageNet, against supervised replay methods (ER, A-GEM, iCaRL, FDR, DER,
DER++) and the supervised-contrastive replay method Co$^{2}$L. In the class-incremental setting, UCLIR (LevelNets) achieves $21.28\%$, above the contrastive baseline Co$^{2}$L ($13.88\%$) and DER++ ($12\%$). UCLIR (SideNets) reaches $12.68\%$, still competitive with DER and DER++. In the task-incremental setting, the supervised and contrastive baselines perform better. UCLIR (LevelNets) reaches $39.59\%$, close to DER ($40.22\%$) and FDR ($40.36\%$), while UCLIR (SideNets) is lower at $28.52\%$. As on the other benchmarks, UCLIR performs best in the class-incremental setting, while the supervised and replay-based methods perform better in the task-incremental setting.

\begin{table*}
\caption{CIL accuracy comparison on 10-Splits Tiny ImageNet.} 
\centering
\small
\setlength{\tabcolsep}{3pt} 
\begin{tabular}{l  l l  c }
\hline
Method & Supervision & Type & ACC.(\%)  \\
\hline
ER \cite{riemerlearning}          & Supervised & Replay & 8.49 $\pm$ 0.16 \\
A-GEM\cite{chaudhry2019efficient}       & Supervised & Replay & 8.07 $\pm$ 0.08 \\
iCARL \cite{rebuffi2017icarl}            & Supervised & Replay + distillation & 7.53 $\pm$ 0.79 \\
FDR \cite{benjaminmeasuring}        & Supervised & Replay & 8.70 $\pm$ 0.19 \\
DER \cite{buzzega2020dark}                & Supervised & Replay & 11.87 $\pm$ 0.78 \\
DER++ \cite{buzzega2020dark}             & Supervised & Replay & 10.96 $\pm$ 1.17 \\
$\mathrm{Co^{2}L}$ \cite{cha2021co2l}               & Contrastive & Replay & 13.88 $\pm$ 0.40 \\
UCLIR (LevelNets)  & Unsupervised & Replay-free & \textbf{21.28 $\pm$ 0.30} \\
UCLIR (SideNets)	  & Unsupervised & Replay-free & 12.68 $\pm$ 0.56 \\  %41.09 (not avg)
%%%%%%%%%%% CIL average: 43.86
\hline
\end{tabular}
\label{table:tiny}
\end{table*}

\begin{table}
\caption{Ablation study on the effect of using multiple representations on CIL accuracy, conducted on 5-Splits CIFAR-10 with LevelNets using L encoders and SideNets using S encoders}\label{tbl1}
\begin{tabular}{l | c }
\hline
 Settings& Accuracy\\

\hline
L1 & 65.53 \\
L1 + L2 & 65.28 \\
L1 + L2 + L3 & \textbf{84.07} \\

\hline
S1  & 47.23 \\
S1 + S2  & 44.90 \\
S1 + S2 + S3 &   52.69\\
S1 + S2 + S3 + S4 &  53.38 \\
S1 + S2 + S3 + S4 + S5 &  \textbf{55.20}\\
\hline
\end{tabular}
\label{table:nets}
\end{table}
\subsection{Ablation study}

\textbf{Effect of the encoders.} Table \ref{table:nets} presents the analysis on the effect of incorporating multiple representations on CIL accuracy. Using a single low-level encoder (L1) in LevelNets shows a performance of 65.53\%, while combining two levels (L1+L2) does not yield improvement and slightly degrades accuracy. However, integrating three hierarchical representations (L1+L2+L3) leads to a performance boost, highlighting the importance of deeper multi-level feature aggregation for resolving class ambiguity. Similar performance observed using SideNets, where the combination of S1,S2, and S2 is insufficient, whereas progressively adding more side representations consistently improves performance. The best result is achieved when all five side representations are used. These results demonstrate the contribution of each level or side representation in the continual learning process.

\begin{figure*}
\centering
\includegraphics[width=1\linewidth]{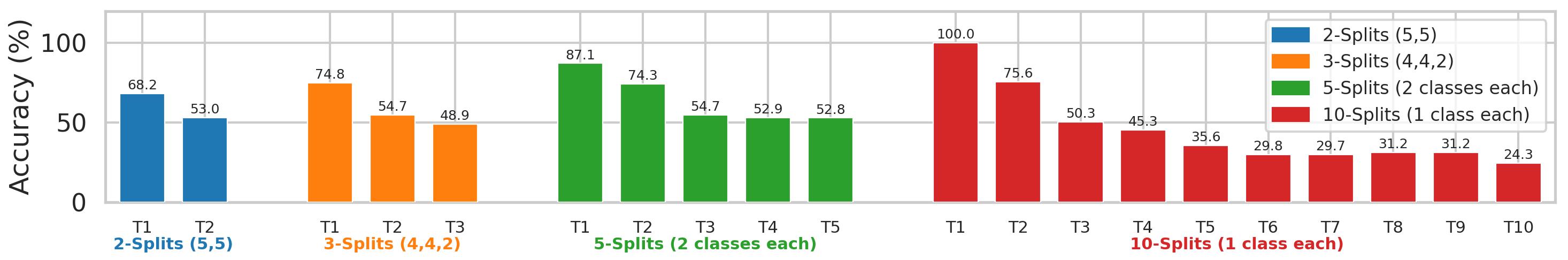}
\caption{Effect of the number of splits on the accuracy under four task-split configurations on CIFAR-10 dataset.}
\label{fig:ab_splits} 
\end{figure*}

\textbf{Effect of the number of splits.} Figure~\ref{fig:ab_splits} shows the per-task accuracy of the model under four
task-split configurations of the same dataset, from  $2$-split ($5{+}5$ classes) to  $10$-split ($1$ class per task). We observe that the starting accuracy on the first task rises as the the number of tasks per-split becomes smaller, from $68.2\%$ for $2$-splits to $100.0\%$ for $10$-splits since each task poses an easier discrimination problem when it contains fewer classes. However, smaller splits also incur a longer incremental sequence and therefore steeper cumulative degradation, like in the case of the $10$-split configuration where the accuracy falls from $100.0\%$ to $24.3\%$ across ten tasks. The figure shows a key trade-off in class-incremental learning, where shorter tasks are easier on their own but having more of them leads to more forgetting overall.

\begin{figure*}
\centering
\includegraphics[width=1\linewidth]{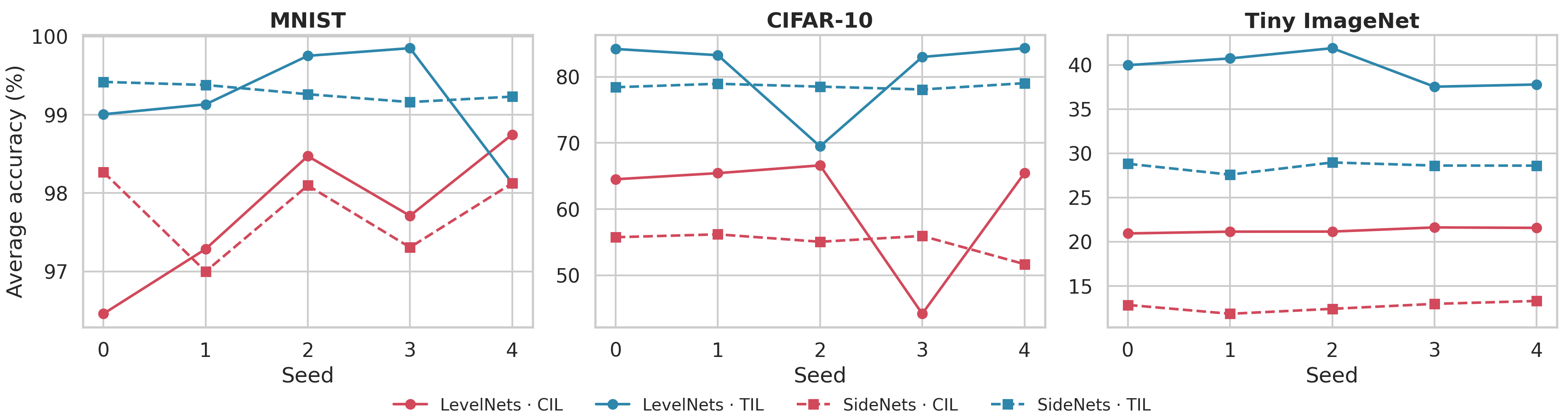}
\caption{Sensitivity of average accuracy across random seeds for LevelNets and SideNets under CIL and TIL.}
\label{fig:seeds} 
\end{figure*}

\textbf{Sensitivity to random seed.} Figure~\ref{fig:seeds} shows each seed's average accuracy for the five runs. On MNIST, the curves stay within a narrow band across the five seeds, where LevelNets CIL moves between about $96.5\%$ and $98.7\%$ while the three other
curves have minor change. Tiny ImageNet is the most consistent of all, since every curve is almost a flat line
across seeds in both LevelNets and SideNets settings, showing that the results here do not depend on the random seed. CIFAR-10 is the exception with LevelNets CIL drops sharply on seed~$3$ to about $44\%$ while staying near $65\%$ on the other four seeds. This single low run is what produces the
wide confidence interval reported for that setting in Table~\ref{tab:results}.
SideNets CIL on the same benchmark stays close to $55\%$ across all seeds. Two observations hold across all three datasets. First, the TIL curves
vary less than the CIL curves (red), which means that task-incremental accuracy is more reproducible than class-incremental accuracy. Second, SideNets is usually steadier than LevelNets, even though LevelNets often reaches the higher average.

\subsection{Statistical analysis}
Table~\ref{tab:results} presents class-incremental (CIL) and task-incremental (TIL) accuracy for LevelNets and SideNets across the three datasets, averaged
over five seeds with standard deviations and $95\%$ confidence intervals. The spread of the results varies considerably by dataset and settings. On MNIST, both LevelNets and SideNets have overlapping confidence intervals in both settings, and their differences are not statistically significant. On CIFAR-10, LevelNets shows much larger variability than SideNets with a CIL standard deviation wider ($\pm 9.56$ vs.\ $\pm 1.86$) and a broad confidence interval, which indicates strong sensitivity to the random seed. TIL interval is similarly wide ($\pm 6.38$ vs.\ $\pm 0.39$). On Tiny ImageNet both LevelNets and SideNets are stable, with small standard deviations and narrow intervals under CIL and TIL. In general, SideNets produces tighter intervals and lower variance than LevelNets across the datasets.

\subsection{Per-task accuracy analysis}

\begin{figure*}
\centering
\includegraphics[width=1\linewidth]{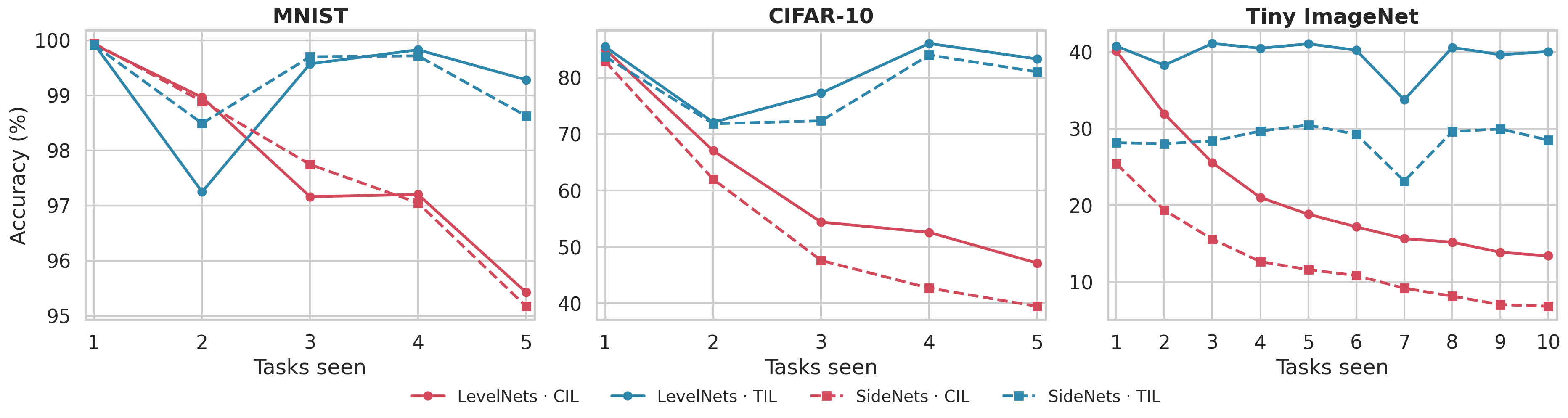}
\caption{Per-task accuracy for LevelNets and SideNets under CIL and TIL.}
\label{fig:tasks} 
\end{figure*}

Figure~\ref{fig:tasks} traces per-task accuracy along the incremental sequence for both LevelNets and SideNets. Under CIL, the curves decline as more
tasks are seen on the three datasets, but the level and the steepness of the decline differ. On MNIST the CIL curves stay above $95\%$, on CIFAR-10 they fall from about $85\%$ to the $40$--$47\%$ range, and on Tiny ImageNet they drop from about $40\%$ to below $15\%$ by the final task. The gap between the two settings also grows with dataset difficulty, from almost no separation on MNIST to a clear and consistent margin for LevelNets on CIFAR-10 and Tiny ImageNet, where the CIL curve stays above SideNets across the whole sequence.

Under TIL, the curves are much flatter on every dataset and stay far higher than the CIL curves. On MNIST they remain close to $99\%$, on CIFAR-10 they hold in the $72$--$85\%$  with only a mild dip in the middle of the sequence, and on Tiny ImageNet they stay near $40\%$ (LevelNets) and $30\%$ (SideNets) across all ten tasks. Overall, in both settings, the class-incremental accuracy degrades as the number of tasks increases, but the TIL curves remain stable across the sequence on every dataset since TIL evaluates only on each task data while CIL evaluates on all classes seen so far at each task.

\begin{table*}[t]
\centering
\small
\setlength{\tabcolsep}{3pt} 
\caption{Class-incremental (CIL) and task-incremental (TIL) accuracy (\%), averaged over 5 seeds. Values are mean $\pm$ sample standard deviation; 95\% confidence intervals (Student's $t$, $n{=}5$) in brackets.}
\label{tab:results}
\begin{tabular}{llcccc}
\toprule
& & \multicolumn{2}{c}{\textbf{CIL}} & \multicolumn{2}{c}{\textbf{TIL}} \\
\cmidrule(lr){3-4}\cmidrule(lr){5-6}
\textbf{Dataset} & \textbf{Encoder} & Mean $\pm$ Std & 95\% CI & Mean $\pm$ Std & 95\% CI \\
\midrule
\multirow{2}{*}{MNIST} & LevelNets & 97.74 $\pm$ 0.92 & [96.59, 98.88] & 99.17 $\pm$ 0.69 & [98.31, 100.03] \\
 & SideNets & 97.76 $\pm$ 0.57 & [97.05, 98.46] & 99.29 $\pm$ 0.11 & [99.16, 99.42] \\
\midrule
\multirow{2}{*}{CIFAR-10} & LevelNets & 61.23 $\pm$ 9.56 & [49.36, 73.10] & 80.85 $\pm$ 6.38 & [72.93, 88.77] \\
 & SideNets & 54.93 $\pm$ 1.86 & [52.62, 57.23] & 78.60 $\pm$ 0.39 & [78.11, 79.08] \\
\midrule
\multirow{2}{*}{Tiny ImageNet} & LevelNets & 21.28 $\pm$ 0.30 & [20.92, 21.65] & 39.59 $\pm$ 1.89 & [37.25, 41.93] \\
 & SideNets & 12.68 $\pm$ 0.56 & [11.98, 13.37] & 28.52 $\pm$ 0.54 & [27.85, 29.20] \\
\bottomrule
\end{tabular}
\end{table*}

\subsection{Results on a sequence of datasets (5-Datasets)}
\begin{figure}
\centering
\includegraphics[width=0.75\linewidth]{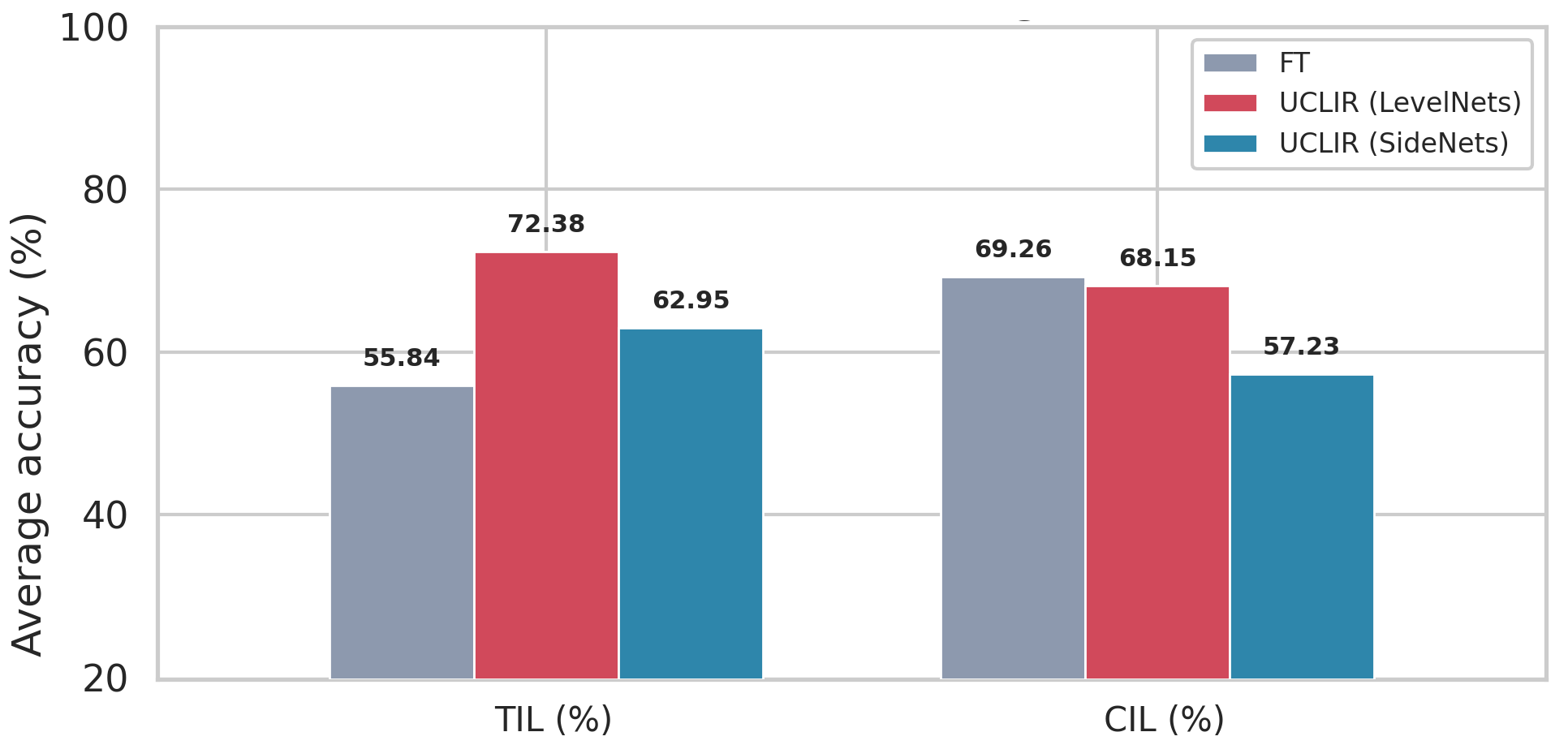}
\caption{Average accuracy over the five tasks for fine-tuning (FT) and UCLIR with LevelNets and SideNets. }
\label{fig:five_compare} 
\end{figure}

Figure~\ref{fig:five_compare} compares the two UCLIR  against a  fine-tuning (FT) baseline on the 5-Datasets, averaged over the five
tasks. In the TIL setting both UCLIR variants clearly surpass FT with LevelNets reaches $72.38\%$ and SideNets $62.95\%$ compared to $55.84\%$ for FT, showing that
independent representation learning substantially reduces forgetting relative to unconstrained sequential training. In the CIL setting LevelNets ($68.15\%$) is essentially on par with FT ($69.26\%$). UCLIR's advantage is most achieved under TIL, where preserving per-task structure matters most, and is competitive under the harder settings of TIL.

\begin{figure}
\centering
\includegraphics[width=0.75\linewidth]{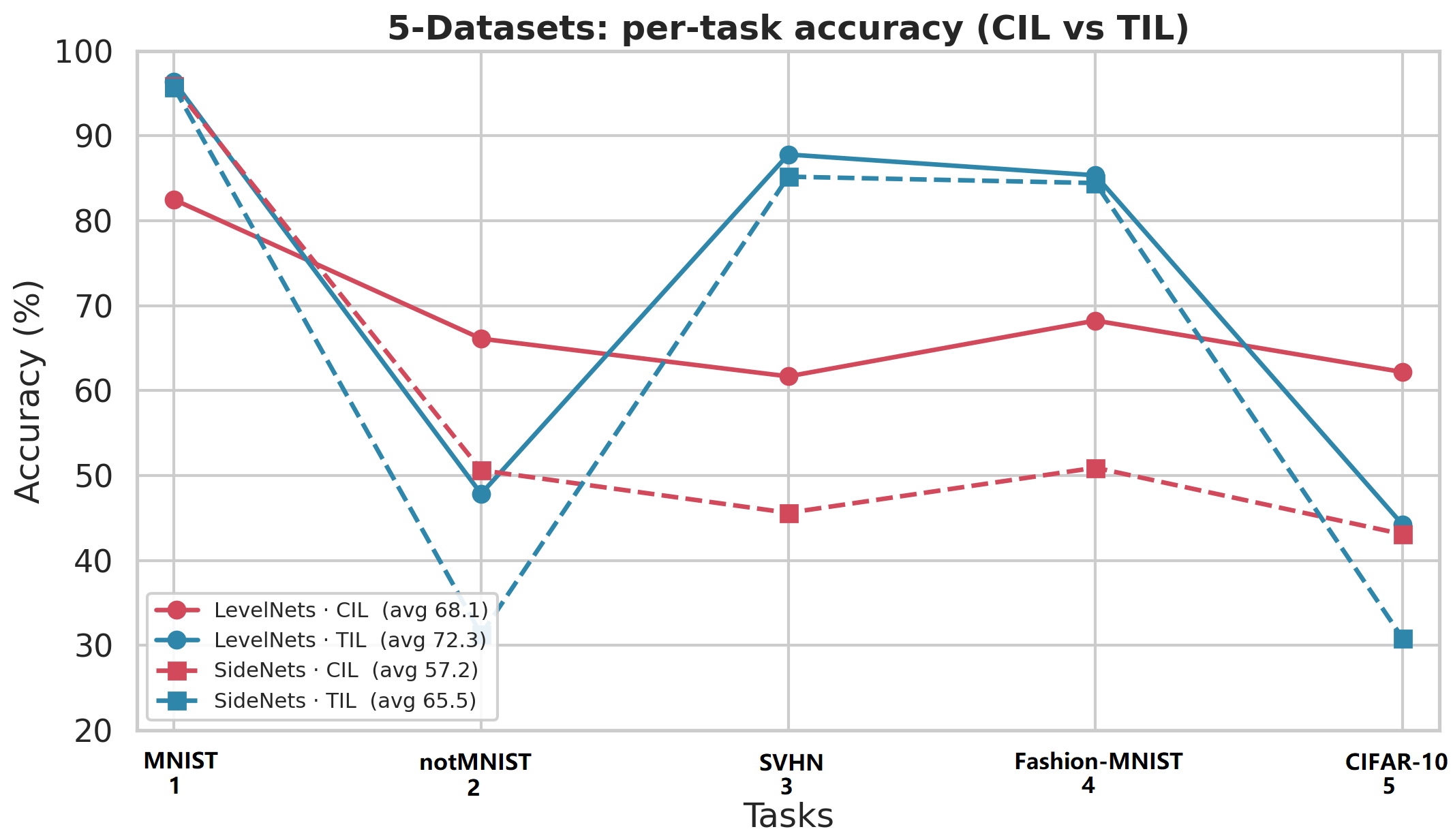}
\caption{Per-task CIL and TIL accuracy
for LevelNets and SideNets across the five sequential datasets.}
\label{fig:five} 
\end{figure}

Figure~\ref{fig:five} shows per-task accuracy across the five sequential datasets. All four curves follow the same shape, which comes from the differing difficulty of the individual datasets rather than from the learning method. The accuracy is high on the easy digit and clothing tasks (MNIST, SVHN, Fashion-MNIST) and drops on the harder tasks (notMNIST  and CIFAR-10). The two TIL curves (blue) rise and fall most sharply. They are the most sensitive to which dataset is being evaluated. However, the CIL curves (red) are flatter, where the accuracy stays in a narrower intervale across the sequence.

\begin{figure*}
\centering
\includegraphics[width=1\linewidth]{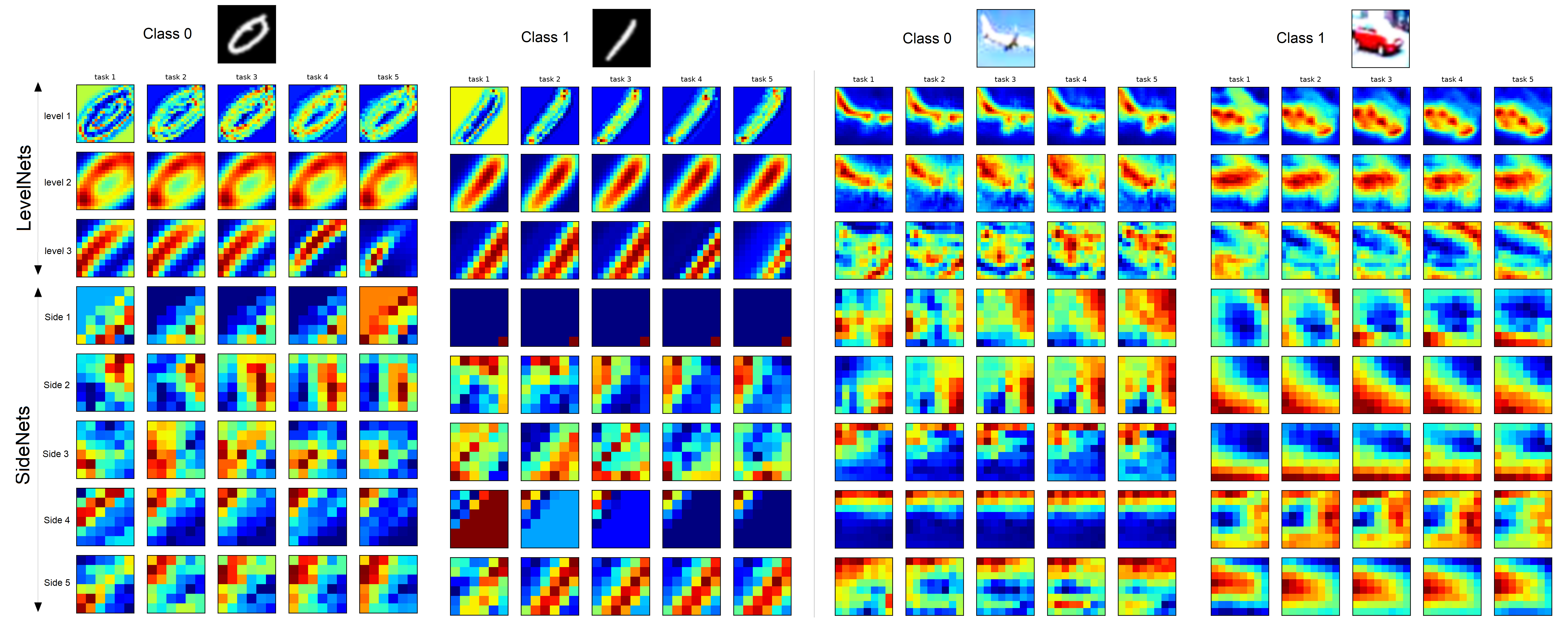}
\caption{Activation maps preservation across tasks for  5-Splits MNIST and 5-Splits CIFAR-10. Classes from task 1.}
\label{fig:feat_mncif} 
\end{figure*}

\begin{figure*}
\centering
\includegraphics[width=1\linewidth]{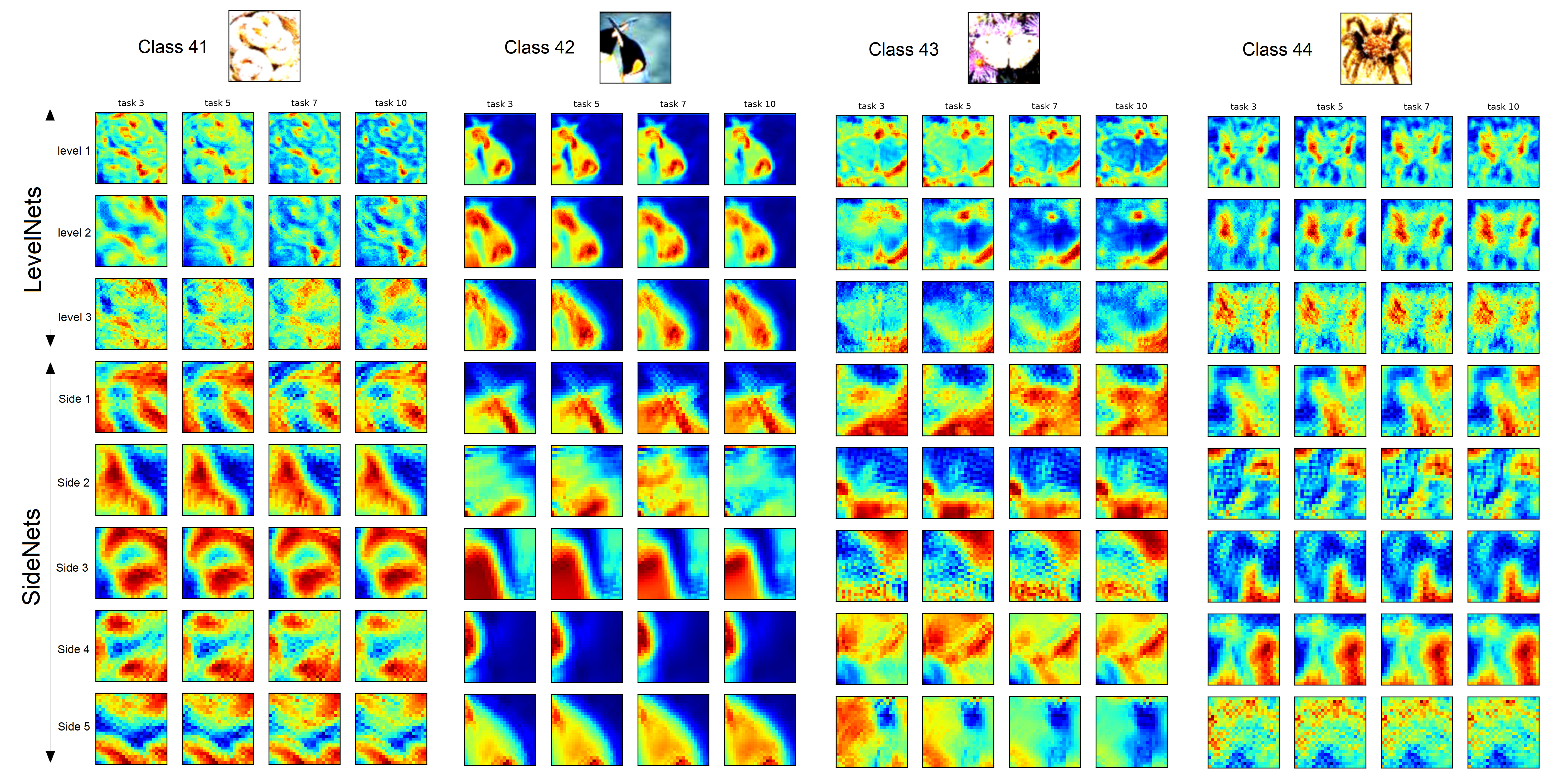}
\caption{Activation maps preservation across tasks for 10-Splits Tiny ImageNet. Classes from task 3.}
\label{fig:feat_tiny} 
\end{figure*}

\begin{figure*}
\centering
\includegraphics[width=1\linewidth]{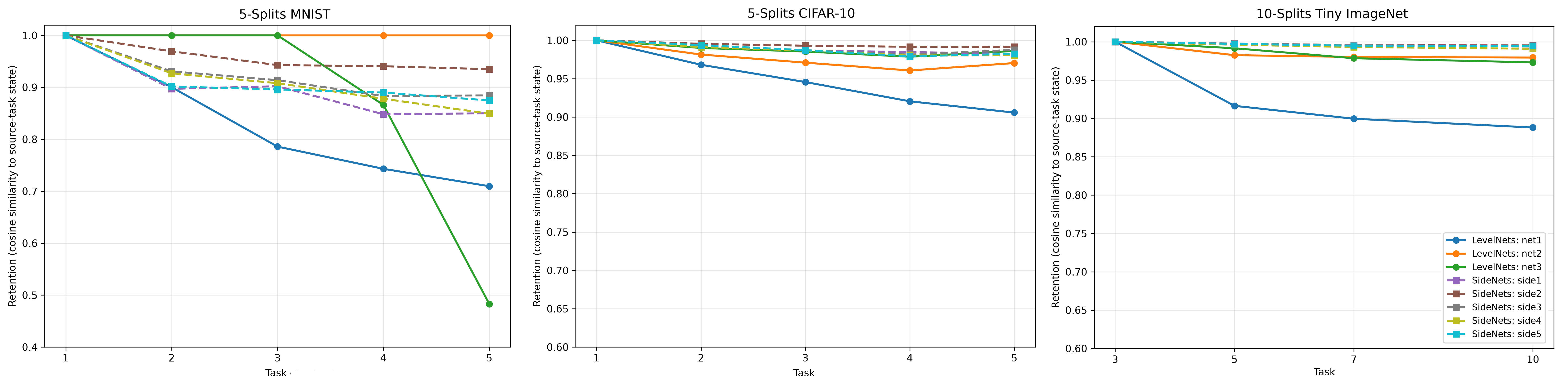}
\caption{Feature retention across tasks calculated as Cosine similarity between the features of the class sample task and the next tasks.}
\label{fig:retention} 
\end{figure*}

\subsection{Visualisation of Feature Retention}

We visualize activation maps (Figure~\ref{fig:feat_mncif} and~\ref{fig:feat_tiny}) to show  qualitatively how the activation pattern of each branch for a fixed example image changes as more tasks are learned. For both settings and across all three datasets, most branches reproduce a visually near identical heatmap from the task at which the example's class was learned through to the following tasks, which indicates  that the corresponding representation is largely undisturbed by subsequent tasks' features. The clearest exception is LevelNets level 3 on MNIST (Figure~\ref{fig:feat_mncif}, class 0), whose activation map is stable from task 1 to task 4 but changes at task 5, from a broad diagonal activation to a small localized pattern. LevelNets level 1 similarly shows a slower, visible drift on both MNIST and Tiny ImageNet, whereas levels 2 and level 3 (except for MNIST drift) and all SideNets sides remain close to their original pattern throughout. Several SideNets branches also reveal patch local specialization directly in the activation maps. For CIFAR-10, despite the heterogeneity of classes compared to MNIST, it shows stability of feature preservation across tasks for both LevelNets and SideNets. This stability is highly confirmed with Tiny ImageNet with more diverse and number of classes.

The feature preservation is quantitatively measured through Cosine similarity between the  features generated by the model's task of the source sample and the following tasks' models. Figure~\ref{fig:retention} shows this measurement across tasks for the samples shown in Figure \ref{fig:feat_mncif} and Figure \ref{fig:feat_tiny}. LevelNets level~1 declines the most consistently of any branch on all three datasets, falling to roughly 0.71 by task~5 on MNIST, 0.91 by task 5 on CIFAR-10, and 0.89 by task 10 on Tiny ImageNet. LevelNets level 3 is the most extreme case. It remains at 1.0 through task 3 and then drops sharply to about 0.48 by task~5 on MNIST, while on CIFAR-10 and Tiny ImageNet the same branch stays close to 0.97--0.98 throughout. This drop reflects the sever change in activation map discusses previously. SideNets branches are the most stable across the tasks and datasets, where the similarity stays  within a narrow band (roughly 0.83--0.93 on MNIST and above 0.97 on CIFAR-10 and Tiny ImageNet) with no severe drops.

\subsection{Conclusion}

This paper presented a replay-free unsupervised continual learning framework that mitigates catastrophic forgetting by learning and regularizing multiple independent complementary representations instead of a single shared representation. Our approach localizes task-specific updates within independent representations, reducing inter-dependencies and harmful interference as the number of tasks increases. We further provided a theoretical analysis showing that independent regularization induces a block diagonal parameter importance geometry, under which the forgetting quadratic separates additively across representations and per-representation EWC controls exactly the surviving interference terms, and we showed that the depth-wise and space-wise structures are two instantiations of this principle. Experiments on 5-Splits MNIST, 5-Datasets, and 5-Splits CIFAR-10  demonstrate that the proposed method achieves competitive or superior performance compared to supervise, contrastive and replay-based methods, particularly under the CIL setting, which reflects real-world continual learning scenarios.

\textbf{Limitations.} Despite these results, our approach has the following limitations. First, maintaining several independent representations increases the
number of parameters and the training cost relative to a single shared encoder, and this cost grows with the number of representations. Second, the two
structural instantiations differ in stability, the depth-wise variant (LevelNets) attains higher mean accuracy but shows  sensitivity to the random seed
on CIFAR-10, whereas the space-wise variant (SideNets) is more consistent but lower on average, so neither dominates across all settings. Finally, our
evaluation is limited to relatively small-scale benchmarks (MNIST, 5-Datasets, CIFAR-10, and Tiny ImageNet) and a fixed number of representations per structure, while the behaviour of the method on larger images, longer task sequences, and with an adaptively chosen number of representations remains to be explored.

\textbf{Future Work.} Several directions follow naturally from the previous limitations. The parameter and compute overhead motivates studying how many representations are actually needed, including mechanisms that add or prune representations adaptively as tasks arrive, so that capacity scales with task
difficulty rather than being fixed in advance. The differing stability between the depth representations and the spatial representations suggests investigating hybrid or learned partitioning that combine the accuracy of the former with the robustness of the latter. Finally, validating the framework on larger and more diverse datasets and on longer task sequences would show whether the method still works well when there are many more tasks and classes.\\

%%===================================================%%
%% For presentation purpose, we have included        %%
%% \bigskip command. Please ignore this.             %%
%%===================================================%%

%%===========================================================================================%%
%% If you are submitting to one of the Nature Portfolio journals, using the eJP submission   %%
%% system, please include the references within the manuscript file itself. You may do this  %%
%% by copying the reference list from your .bbl file, paste it into the main manuscript .tex %%
%% file, and delete the associated \verb+\bibliography+ commands.                            %%
%%===========================================================================================%%

\bibliography{biblio}% common bib file

%% BioMed_Central_Bib_Style_v1.01

\begin{thebibliography}{31}
% BibTex style file: bmc-mathphys.bst (version 2.1), 2014-07-24
\ifx \bisbn   \undefined \def \bisbn  #1{ISBN #1}\fi
\ifx \binits  \undefined \def \binits#1{#1}\fi
\ifx \bauthor  \undefined \def \bauthor#1{#1}\fi
\ifx \batitle  \undefined \def \batitle#1{#1}\fi
\ifx \bjtitle  \undefined \def \bjtitle#1{#1}\fi
\ifx \bvolume  \undefined \def \bvolume#1{\textbf{#1}}\fi
\ifx \byear  \undefined \def \byear#1{#1}\fi
\ifx \bissue  \undefined \def \bissue#1{#1}\fi
\ifx \bfpage  \undefined \def \bfpage#1{#1}\fi
\ifx \blpage  \undefined \def \blpage #1{#1}\fi
\ifx \burl  \undefined \def \burl#1{\textsf{#1}}\fi
\ifx \doiurl  \undefined \def \doiurl#1{\url{https://doi.org/#1}}\fi
\ifx \betal  \undefined \def \betal{\textit{et al.}}\fi
\ifx \binstitute  \undefined \def \binstitute#1{#1}\fi
\ifx \binstitutionaled  \undefined \def \binstitutionaled#1{#1}\fi
\ifx \bctitle  \undefined \def \bctitle#1{#1}\fi
\ifx \beditor  \undefined \def \beditor#1{#1}\fi
\ifx \bpublisher  \undefined \def \bpublisher#1{#1}\fi
\ifx \bbtitle  \undefined \def \bbtitle#1{#1}\fi
\ifx \bedition  \undefined \def \bedition#1{#1}\fi
\ifx \bseriesno  \undefined \def \bseriesno#1{#1}\fi
\ifx \blocation  \undefined \def \blocation#1{#1}\fi
\ifx \bsertitle  \undefined \def \bsertitle#1{#1}\fi
\ifx \bsnm \undefined \def \bsnm#1{#1}\fi
\ifx \bsuffix \undefined \def \bsuffix#1{#1}\fi
\ifx \bparticle \undefined \def \bparticle#1{#1}\fi
\ifx \barticle \undefined \def \barticle#1{#1}\fi
\bibcommenthead
\ifx \bconfdate \undefined \def \bconfdate #1{#1}\fi
\ifx \botherref \undefined \def \botherref #1{#1}\fi
\ifx \url \undefined \def \url#1{\textsf{#1}}\fi
\ifx \bchapter \undefined \def \bchapter#1{#1}\fi
\ifx \bbook \undefined \def \bbook#1{#1}\fi
\ifx \bcomment \undefined \def \bcomment#1{#1}\fi
\ifx \oauthor \undefined \def \oauthor#1{#1}\fi
\ifx \citeauthoryear \undefined \def \citeauthoryear#1{#1}\fi
\ifx \endbibitem  \undefined \def \endbibitem {}\fi
\ifx \bconflocation  \undefined \def \bconflocation#1{#1}\fi
\ifx \arxivurl  \undefined \def \arxivurl#1{\textsf{#1}}\fi
\csname PreBibitemsHook\endcsname

%%% 1
\bibitem[\protect\citeauthoryear{Shin et~al.}{2017}]{shin2017continual}
\begin{botherref}
\oauthor{\bsnm{Shin}, \binits{H.}},
\oauthor{\bsnm{Lee}, \binits{J.K.}},
\oauthor{\bsnm{Kim}, \binits{J.}},
\oauthor{\bsnm{Kim}, \binits{J.}}:
Continual learning with deep generative replay.
Advances in neural information processing systems
\textbf{30}
(2017)
\end{botherref}
\endbibitem

%%% 2
\bibitem[\protect\citeauthoryear{Rusu et~al.}{2016}]{rusu2016progressive}
\begin{botherref}
\oauthor{\bsnm{Rusu}, \binits{A.A.}},
\oauthor{\bsnm{Rabinowitz}, \binits{N.C.}},
\oauthor{\bsnm{Desjardins}, \binits{G.}},
\oauthor{\bsnm{Soyer}, \binits{H.}},
\oauthor{\bsnm{Kirkpatrick}, \binits{J.}},
\oauthor{\bsnm{Kavukcuoglu}, \binits{K.}},
\oauthor{\bsnm{Pascanu}, \binits{R.}},
\oauthor{\bsnm{Hadsell}, \binits{R.}}:
Progressive neural networks.
arXiv preprint arXiv:1606.04671
(2016)
\end{botherref}
\endbibitem

%%% 3
\bibitem[\protect\citeauthoryear{Kirkpatrick et~al.}{2017}]{kirkpatrick2017overcoming}
\begin{barticle}
\bauthor{\bsnm{Kirkpatrick}, \binits{J.}},
\bauthor{\bsnm{Pascanu}, \binits{R.}},
\bauthor{\bsnm{Rabinowitz}, \binits{N.}},
\bauthor{\bsnm{Veness}, \binits{J.}},
\bauthor{\bsnm{Desjardins}, \binits{G.}},
\bauthor{\bsnm{Rusu}, \binits{A.A.}},
\bauthor{\bsnm{Milan}, \binits{K.}},
\bauthor{\bsnm{Quan}, \binits{J.}},
\bauthor{\bsnm{Ramalho}, \binits{T.}},
\bauthor{\bsnm{Grabska-Barwinska}, \binits{A.}}, \betal:
\batitle{Overcoming catastrophic forgetting in neural networks}.
\bjtitle{Proceedings of the national academy of sciences}
\bvolume{114}(\bissue{13}),
\bfpage{3521}--\blpage{3526}
(\byear{2017})
\end{barticle}
\endbibitem

%%% 4
\bibitem[\protect\citeauthoryear{Yoon et~al.}{2018}]{yoon2018lifelong}
\begin{bchapter}
\bauthor{\bsnm{Yoon}, \binits{J.}},
\bauthor{\bsnm{Yang}, \binits{E.}},
\bauthor{\bsnm{Lee}, \binits{J.}},
\bauthor{\bsnm{Hwang}, \binits{S.J.}}:
\bctitle{Lifelong learning with dynamically expandable networks}.
In: \bbtitle{International Conference on Learning Representations}
(\byear{2018})
\end{bchapter}
\endbibitem

%%% 5
\bibitem[\protect\citeauthoryear{Hebb}{1949}]{hebb1949organization}
\begin{bbook}
\bauthor{\bsnm{Hebb}, \binits{D.O.}}:
\bbtitle{The Organization of Behavior: A Neuropsychological Theory}.
\bpublisher{Wiley}, \blocation{???}
(\byear{1949})
\end{bbook}
\endbibitem

%%% 6
\bibitem[\protect\citeauthoryear{Bi and Poo}{1998}]{bi_synaptic_1998-1}
\begin{barticle}
\bauthor{\bsnm{Bi}, \binits{G.-q.}},
\bauthor{\bsnm{Poo}, \binits{M.-m.}}:
\batitle{Synaptic modifications in cultured hippocampal neurons: Dependence on spike timing, synaptic strength, and postsynaptic cell type}.
\bjtitle{The Journal of Neuroscience}
\bvolume{18}(\bissue{24}),
\bfpage{10464}--\blpage{10472}
(\byear{1998})
\doiurl{10.1523/JNEUROSCI.18-24-10464.1998}
\end{barticle}
\endbibitem

%%% 7
\bibitem[\protect\citeauthoryear{Ahn et~al.}{2019}]{ahn2019uncertainty}
\begin{botherref}
\oauthor{\bsnm{Ahn}, \binits{H.}},
\oauthor{\bsnm{Cha}, \binits{S.}},
\oauthor{\bsnm{Lee}, \binits{D.}},
\oauthor{\bsnm{Moon}, \binits{T.}}:
Uncertainty-based continual learning with adaptive regularization.
Advances in neural information processing systems
\textbf{32}
(2019)
\end{botherref}
\endbibitem

%%% 8
\bibitem[\protect\citeauthoryear{Chen et~al.}{2021}]{chen2021overcoming}
\begin{bchapter}
\bauthor{\bsnm{Chen}, \binits{P.-H.}},
\bauthor{\bsnm{Wei}, \binits{W.}},
\bauthor{\bsnm{Hsieh}, \binits{C.-J.}},
\bauthor{\bsnm{Dai}, \binits{B.}}:
\bctitle{Overcoming catastrophic forgetting by bayesian generative regularization}.
In: \bbtitle{International Conference on Machine Learning},
pp. \bfpage{1760}--\blpage{1770}
(\byear{2021}).
\bcomment{PMLR}
\end{bchapter}
\endbibitem

%%% 9
\bibitem[\protect\citeauthoryear{Ercsey-Ravasz et~al.}{2013}]{ercsey2013predictive}
\begin{barticle}
\bauthor{\bsnm{Ercsey-Ravasz}, \binits{M.}},
\bauthor{\bsnm{Markov}, \binits{N.T.}},
\bauthor{\bsnm{Lamy}, \binits{C.}},
\bauthor{\bsnm{Van~Essen}, \binits{D.C.}},
\bauthor{\bsnm{Knoblauch}, \binits{K.}},
\bauthor{\bsnm{Toroczkai}, \binits{Z.}},
\bauthor{\bsnm{Kennedy}, \binits{H.}}:
\batitle{A predictive network model of cerebral cortical connectivity based on a distance rule}.
\bjtitle{Neuron}
\bvolume{80}(\bissue{1}),
\bfpage{184}--\blpage{197}
(\byear{2013})
\end{barticle}
\endbibitem

%%% 10
\bibitem[\protect\citeauthoryear{Kobatake and Tanaka}{1994}]{kobatake1994neuronal}
\begin{barticle}
\bauthor{\bsnm{Kobatake}, \binits{E.}},
\bauthor{\bsnm{Tanaka}, \binits{K.}}:
\batitle{Neuronal selectivities to complex object features in the ventral visual pathway of the macaque cerebral cortex}.
\bjtitle{Journal of neurophysiology}
\bvolume{71}(\bissue{3}),
\bfpage{856}--\blpage{867}
(\byear{1994})
\end{barticle}
\endbibitem

%%% 11
\bibitem[\protect\citeauthoryear{Zenke et~al.}{2017}]{zenke2017continual}
\begin{bchapter}
\bauthor{\bsnm{Zenke}, \binits{F.}},
\bauthor{\bsnm{Poole}, \binits{B.}},
\bauthor{\bsnm{Ganguli}, \binits{S.}}:
\bctitle{Continual learning through synaptic intelligence}.
In: \bbtitle{International Conference on Machine Learning},
pp. \bfpage{3987}--\blpage{3995}
(\byear{2017}).
\bcomment{PMLR}
\end{bchapter}
\endbibitem

%%% 12
\bibitem[\protect\citeauthoryear{Lopez-Paz and Ranzato}{2017}]{lopez2017gradient}
\begin{botherref}
\oauthor{\bsnm{Lopez-Paz}, \binits{D.}},
\oauthor{\bsnm{Ranzato}, \binits{M.}}:
Gradient episodic memory for continual learning.
Advances in neural information processing systems
\textbf{30}
(2017)
\end{botherref}
\endbibitem

%%% 13
\bibitem[\protect\citeauthoryear{Ebrahimi et~al.}{2020}]{ebrahimi2020adversarial}
\begin{bchapter}
\bauthor{\bsnm{Ebrahimi}, \binits{S.}},
\bauthor{\bsnm{Meier}, \binits{F.}},
\bauthor{\bsnm{Calandra}, \binits{R.}},
\bauthor{\bsnm{Darrell}, \binits{T.}},
\bauthor{\bsnm{Rohrbach}, \binits{M.}}:
\bctitle{Adversarial continual learning}.
In: \bbtitle{European Conference on Computer Vision},
pp. \bfpage{386}--\blpage{402}
(\byear{2020}).
\bcomment{Springer}
\end{bchapter}
\endbibitem

%%% 14
\bibitem[\protect\citeauthoryear{Kozal and Wozniak}{2023}]{kozal2023increasing}
\begin{barticle}
\bauthor{\bsnm{Kozal}, \binits{J.}},
\bauthor{\bsnm{Wozniak}, \binits{M.}}:
\batitle{Increasing depth of neural networks for life-long learning}.
\bjtitle{Information Fusion}
\bvolume{98},
\bfpage{101829}
(\byear{2023})
\end{barticle}
\endbibitem

%%% 15
\bibitem[\protect\citeauthoryear{Zbontar et~al.}{2021}]{zbontar2021barlow}
\begin{bchapter}
\bauthor{\bsnm{Zbontar}, \binits{J.}},
\bauthor{\bsnm{Jing}, \binits{L.}},
\bauthor{\bsnm{Misra}, \binits{I.}},
\bauthor{\bsnm{LeCun}, \binits{Y.}},
\bauthor{\bsnm{Deny}, \binits{S.}}:
\bctitle{Barlow twins: Self-supervised learning via redundancy reduction}.
In: \bbtitle{International Conference on Machine Learning},
pp. \bfpage{12310}--\blpage{12320}
(\byear{2021}).
\bcomment{PMLR}
\end{bchapter}
\endbibitem

%%% 16
\bibitem[\protect\citeauthoryear{Chen et~al.}{2020}]{chen2020simple}
\begin{bchapter}
\bauthor{\bsnm{Chen}, \binits{T.}},
\bauthor{\bsnm{Kornblith}, \binits{S.}},
\bauthor{\bsnm{Norouzi}, \binits{M.}},
\bauthor{\bsnm{Hinton}, \binits{G.}}:
\bctitle{A simple framework for contrastive learning of visual representations}.
In: \bbtitle{International Conference on Machine Learning},
pp. \bfpage{1597}--\blpage{1607}
(\byear{2020}).
\bcomment{PmLR}
\end{bchapter}
\endbibitem

%%% 17
\bibitem[\protect\citeauthoryear{Cha et~al.}{2021}]{cha2021co2l}
\begin{bchapter}
\bauthor{\bsnm{Cha}, \binits{H.}},
\bauthor{\bsnm{Lee}, \binits{J.}},
\bauthor{\bsnm{Shin}, \binits{J.}}:
\bctitle{Co2l: Contrastive continual learning}.
In: \bbtitle{Proceedings of the IEEE/CVF International Conference on Computer Vision},
pp. \bfpage{9516}--\blpage{9525}
(\byear{2021})
\end{bchapter}
\endbibitem

%%% 18
\bibitem[\protect\citeauthoryear{Fini et~al.}{2022}]{fini2022self}
\begin{bchapter}
\bauthor{\bsnm{Fini}, \binits{E.}},
\bauthor{\bsnm{Da~Costa}, \binits{V.G.T.}},
\bauthor{\bsnm{Alameda-Pineda}, \binits{X.}},
\bauthor{\bsnm{Ricci}, \binits{E.}},
\bauthor{\bsnm{Alahari}, \binits{K.}},
\bauthor{\bsnm{Mairal}, \binits{J.}}:
\bctitle{Self-supervised models are continual learners}.
In: \bbtitle{Proceedings of the IEEE/CVF Conference on Computer Vision and Pattern Recognition},
pp. \bfpage{9621}--\blpage{9630}
(\byear{2022})
\end{bchapter}
\endbibitem

%%% 19
\bibitem[\protect\citeauthoryear{Wen et~al.}{2024}]{wen2024provable}
\begin{bchapter}
\bauthor{\bsnm{Wen}, \binits{Y.}},
\bauthor{\bsnm{Tan}, \binits{Z.}},
\bauthor{\bsnm{Zheng}, \binits{K.}},
\bauthor{\bsnm{Xie}, \binits{C.}},
\bauthor{\bsnm{Huang}, \binits{W.}}:
\bctitle{Provable contrastive continual learning}.
In: \bbtitle{International Conference on Machine Learning},
pp. \bfpage{52819}--\blpage{52838}
(\byear{2024}).
\bcomment{PMLR}
\end{bchapter}
\endbibitem

%%% 20
\bibitem[\protect\citeauthoryear{Cheng et~al.}{2023}]{cheng2023cucl}
\begin{bchapter}
\bauthor{\bsnm{Cheng}, \binits{C.}},
\bauthor{\bsnm{Song}, \binits{J.}},
\bauthor{\bsnm{Zhu}, \binits{X.}},
\bauthor{\bsnm{Zhu}, \binits{J.}},
\bauthor{\bsnm{Gao}, \binits{L.}},
\bauthor{\bsnm{Shen}, \binits{H.}}:
\bctitle{Cucl: Codebook for unsupervised continual learning}.
In: \bbtitle{Proceedings of the 31st ACM International Conference on Multimedia},
pp. \bfpage{1729}--\blpage{1737}
(\byear{2023})
\end{bchapter}
\endbibitem

%%% 21
\bibitem[\protect\citeauthoryear{Cheng et~al.}{2024}]{cheng2024must}
\begin{bchapter}
\bauthor{\bsnm{Cheng}, \binits{H.}},
\bauthor{\bsnm{Wen}, \binits{H.}},
\bauthor{\bsnm{Qiu}, \binits{H.}},
\bauthor{\bsnm{Wang}, \binits{L.}},
\bauthor{\bsnm{Zhang}, \binits{M.}},
\bauthor{\bsnm{Li}, \binits{H.}}:
\bctitle{Must unsupervised continual learning relies on previous information?}
In: \bbtitle{Proceedings of the IEEE/CVF Conference on Computer Vision and Pattern Recognition},
pp. \bfpage{5519}--\blpage{5529}
(\byear{2024})
\end{bchapter}
\endbibitem

%%% 22
\bibitem[\protect\citeauthoryear{Aljundi et~al.}{2018}]{aljundi2018memory}
\begin{bchapter}
\bauthor{\bsnm{Aljundi}, \binits{R.}},
\bauthor{\bsnm{Babiloni}, \binits{F.}},
\bauthor{\bsnm{Elhoseiny}, \binits{M.}},
\bauthor{\bsnm{Rohrbach}, \binits{M.}},
\bauthor{\bsnm{Tuytelaars}, \binits{T.}}:
\bctitle{Memory aware synapses: Learning what (not) to forget}.
In: \bbtitle{Proceedings of the European Conference on Computer Vision (ECCV)},
pp. \bfpage{139}--\blpage{154}
(\byear{2018})
\end{bchapter}
\endbibitem

%%% 23
\bibitem[\protect\citeauthoryear{Li and Hoiem}{2017}]{li2017learning}
\begin{barticle}
\bauthor{\bsnm{Li}, \binits{Z.}},
\bauthor{\bsnm{Hoiem}, \binits{D.}}:
\batitle{Learning without forgetting}.
\bjtitle{IEEE transactions on pattern analysis and machine intelligence}
\bvolume{40}(\bissue{12}),
\bfpage{2935}--\blpage{2947}
(\byear{2017})
\end{barticle}
\endbibitem

%%% 24
\bibitem[\protect\citeauthoryear{Rebuffi et~al.}{2017}]{rebuffi2017icarl}
\begin{bchapter}
\bauthor{\bsnm{Rebuffi}, \binits{S.-A.}},
\bauthor{\bsnm{Kolesnikov}, \binits{A.}},
\bauthor{\bsnm{Sperl}, \binits{G.}},
\bauthor{\bsnm{Lampert}, \binits{C.H.}}:
\bctitle{icarl: Incremental classifier and representation learning}.
In: \bbtitle{Proceedings of the IEEE Conference on Computer Vision and Pattern Recognition},
pp. \bfpage{2001}--\blpage{2010}
(\byear{2017})
\end{bchapter}
\endbibitem

%%% 25
\bibitem[\protect\citeauthoryear{Rao et~al.}{2019}]{rao2019continual}
\begin{botherref}
\oauthor{\bsnm{Rao}, \binits{D.}},
\oauthor{\bsnm{Visin}, \binits{F.}},
\oauthor{\bsnm{Rusu}, \binits{A.}},
\oauthor{\bsnm{Pascanu}, \binits{R.}},
\oauthor{\bsnm{Teh}, \binits{Y.W.}},
\oauthor{\bsnm{Hadsell}, \binits{R.}}:
Continual unsupervised representation learning.
Advances in neural information processing systems
\textbf{32}
(2019)
\end{botherref}
\endbibitem

%%% 26
\bibitem[\protect\citeauthoryear{Riemer et~al.}{2019}]{riemerlearning}
\begin{bchapter}
\bauthor{\bsnm{Riemer}, \binits{M.}},
\bauthor{\bsnm{Cases}, \binits{I.}},
\bauthor{\bsnm{Ajemian}, \binits{R.}},
\bauthor{\bsnm{Liu}, \binits{M.}},
\bauthor{\bsnm{Rish}, \binits{I.}},
\bauthor{\bsnm{Tu}, \binits{Y.}},
\bauthor{\bsnm{Tesauro}, \binits{G.}}:
\bctitle{Learning to learn without forgetting by maximizing transfer and minimizing interference}.
In: \bbtitle{International Conference on Learning Representations}
(\byear{2019})
\end{bchapter}
\endbibitem

%%% 27
\bibitem[\protect\citeauthoryear{Chaudhry et~al.}{2019}]{chaudhry2019efficient}
\begin{bchapter}
\bauthor{\bsnm{Chaudhry}, \binits{A.}},
\bauthor{\bsnm{Marc'Aurelio}, \binits{R.}},
\bauthor{\bsnm{Rohrbach}, \binits{M.}},
\bauthor{\bsnm{Elhoseiny}, \binits{M.}}:
\bctitle{Efficient lifelong learning with a-gem}.
In: \bbtitle{7th International Conference on Learning Representations}
(\byear{2019})
\end{bchapter}
\endbibitem

%%% 28
\bibitem[\protect\citeauthoryear{Benjamin et~al.}{2019}]{benjaminmeasuring}
\begin{bchapter}
\bauthor{\bsnm{Benjamin}, \binits{A.}},
\bauthor{\bsnm{Rolnick}, \binits{D.}},
\bauthor{\bsnm{Kording}, \binits{K.}}:
\bctitle{Measuring and regularizing networks in function space}.
In: \bbtitle{International Conference on Learning Representations}
(\byear{2019})
\end{bchapter}
\endbibitem

%%% 29
\bibitem[\protect\citeauthoryear{Aljundi et~al.}{2019}]{aljundi2019gradient}
\begin{botherref}
\oauthor{\bsnm{Aljundi}, \binits{R.}},
\oauthor{\bsnm{Lin}, \binits{M.}},
\oauthor{\bsnm{Goujaud}, \binits{B.}},
\oauthor{\bsnm{Bengio}, \binits{Y.}}:
Gradient based sample selection for online continual learning.
Advances in neural information processing systems
\textbf{32}
(2019)
\end{botherref}
\endbibitem

%%% 30
\bibitem[\protect\citeauthoryear{Chaudhry et~al.}{2021}]{chaudhry2021using}
\begin{bchapter}
\bauthor{\bsnm{Chaudhry}, \binits{A.}},
\bauthor{\bsnm{Gordo}, \binits{A.}},
\bauthor{\bsnm{Dokania}, \binits{P.}},
\bauthor{\bsnm{Torr}, \binits{P.}},
\bauthor{\bsnm{Lopez-Paz}, \binits{D.}}:
\bctitle{Using hindsight to anchor past knowledge in continual learning}.
In: \bbtitle{Proceedings of the AAAI Conference on Artificial Intelligence},
vol. \bseriesno{35},
pp. \bfpage{6993}--\blpage{7001}
(\byear{2021})
\end{bchapter}
\endbibitem

%%% 31
\bibitem[\protect\citeauthoryear{Buzzega et~al.}{2020}]{buzzega2020dark}
\begin{barticle}
\bauthor{\bsnm{Buzzega}, \binits{P.}},
\bauthor{\bsnm{Boschini}, \binits{M.}},
\bauthor{\bsnm{Porrello}, \binits{A.}},
\bauthor{\bsnm{Abati}, \binits{D.}},
\bauthor{\bsnm{Calderara}, \binits{S.}}:
\batitle{Dark experience for general continual learning: a strong, simple baseline}.
\bjtitle{Advances in neural information processing systems}
\bvolume{33},
\bfpage{15920}--\blpage{15930}
(\byear{2020})
\end{barticle}
\endbibitem

\end{thebibliography}
%% if required, the content of .bbl file can be included here once bbl is generated
%%\input sn-article.bbl

\end{document}